\documentclass[sigconf]{acmart}
\AtBeginDocument{%
  \providecommand\BibTeX{{%
    \normalfont B\kern-0.5em{\scshape i\kern-0.25em b}\kern-0.8em\TeX}}}

\newcommand{\red}[1]{#1}
\newcommand{\blue}[1]{#1}

\newcommand{\green}[1]{#1}

\usepackage{subfigure}
\usepackage{enumitem}
\usepackage{bm}
\usepackage[ruled,vlined]{algorithm2e} %
\usepackage[switch]{lineno}
\usepackage{multicol}

\copyrightyear{2023}
\acmYear{2023}
\setcopyright{acmlicensed}
\acmConference[KDD '23] {Proceedings of the 29th ACM SIGKDD Conference on Knowledge Discovery and Data Mining}{August 6--10, 2023}{Long Beach, CA, USA.}
\acmBooktitle{Proceedings of the 29th ACM SIGKDD Conference on Knowledge Discovery and Data Mining (KDD '23), August 6--10, 2023, Long Beach, CA, USA}
\acmPrice{15.00}
\acmISBN{979-8-4007-0103-0/23/08}
\acmDOI{10.1145/3580305.3599770}
\begin{document}

\title{AdSEE: Investigating the Impact of Image Style Editing on Advertisement Attractiveness}

\author{Liyao Jiang}
\affiliation{
  \institution{University of Alberta}
  \city{Edmonton}
  \state{AB}
  \country{Canada}
}
\email{liyao1@ualberta.ca}

\author{Chenglin Li}
\affiliation{
  \institution{University of Alberta}
  \city{Edmonton}
  \state{AB}
  \country{Canada}
}
\email{ch11@ualberta.ca}

\author{Haolan Chen}
\affiliation{
  \institution{Platform and Content Group, Tencent}
  \city{Shenzhen}
  \country{China}
}
\email{haolanchen@tencent.com}

\author{Xiaodong Gao}
\author{Xinwang Zhong}
\affiliation{
  \institution{Platform and Content Group, Tencent}
  \city{Shenzhen}
  \country{China}
}
\email{cshiudawn@gmail.com}
\email{visionzhong@tencent.com}

\author{Yang Qiu}
\author{Shani Ye}
\affiliation{
  \institution{Platform and Content Group, Tencent}
  \city{Shenzhen}
  \country{China}
}
\email{rickyqqiu@tencent.com}
\email{lisaniye@tencent.com}

\author{Di Niu}
\affiliation{
  \institution{University of Alberta}
  \city{Edmonton}
  \state{AB}
  \country{Canada}
}
\email{dniu@ualberta.ca}

\def \authors{Liyao Jiang, Chenglin Li, Haolan Chen, Xiaodong Gao, Xinwang Zhong, Yang Qiu, Shani Ye, and Di Niu}
\renewcommand{\shortauthors}{Liyao Jiang et al.}

\begin{abstract}
Online advertisements are important elements in e-commerce sites, social media platforms, and search engines. With the increasing popularity of mobile browsing, many online ads are displayed with visual information in the form of a cover image in addition to text descriptions to grab the attention of users. Various recent studies have focused on predicting the click rates of online advertisements aware of visual features or composing optimal advertisement elements to enhance visibility. In this paper, we propose Advertisement Style Editing and Attractiveness Enhancement (AdSEE), which explores whether semantic editing to ads images can affect or alter the popularity of online advertisements. We introduce StyleGAN-based facial semantic editing and inversion to ads images and train a click rate predictor attributing GAN-based face latent representations in addition to traditional visual and textual features to click rates. Through a large collected dataset named QQ-AD, containing 20,527 online ads, we perform extensive offline tests to study how different semantic directions and their edit coefficients may impact click rates. We further design a Genetic Advertisement Editor to efficiently search for the optimal edit directions and intensity given an input ad cover image to enhance its projected click rates. Online A/B tests performed over a period of 5 days have verified the increased click-through rates of AdSEE-edited samples as compared to a control group of original ads, verifying the relation between image styles and ad popularity. We open source the code for AdSEE research at https://github.com/LiyaoJiang1998/adsee.
\end{abstract}


\maketitle

\section{Introduction}
\label{sec:intro}
Online or digital advertisements are crucial elements in e-commerce sites, social media platforms, and search engines. With the increasing popularity of mobile browsing, many online \textit{ads} are displayed on cellphones with visual information frequently in the form of a cover image in addition to text description, since visual information is not only more direct but can also grab people's attention compared to text-only \textit{ads}. In fact, previous studies~\cite{azimi2012impact,cheng2012multimedia} have shown that appealing cover images lead to a higher Click-Through Rate (CTR) in online \textit{ads}. 

Therefore, a number of recent studies on online \textit{ads} have focused on extracting visual features for visual-aware CTR prediction~\cite{chen2016deep,he2016vbpr,cscnn}. 
Furthermore, while many online \textit{ad} images contain human faces, previous studies~\cite{adil2018face, guido2019effects, nasiri2020effect} have verified that incorporating human faces in online \textit{ad} correlates to more attention towards the \textit{ads}, as well as that eye gaze directions have an impact on user response.
Another related research direction focuses on Advertisement Creatives selection~\cite{chen2021automated}, which searches from a large pool of creative elements and templates to compose a good \textit{ad} design. Thanks to recent advancements in generative adversarial networks (GANs), e.g., SytleGAN \cite{stylegan, stylegan2, stylegan3}, image editing has been made possible, especially with respect to facial semantics. However, existing studies have not investigated the impact of style editing in recommender systems. %

In this paper, we propose the Advertisement Style Editing and Attractiveness Enhancement (AdSEE) system, which aims to do a reality check to answer a long-standing question in AI ethics--whether editing the facial style in an online \textit{ad} can enhance its attractiveness? %
AdSEE consists of two parts: 1) a Click Rate Predictor (CRP) to predict the averaged click rate (CR) for any given \textit{ad} in an \textit{ad} category, based on its cover image and text information, and 2) a Genetic Advertisement Editor (GADE) to search for the optimal face editing dimensions and directions, e.g., smile, eye gaze direction, as well as the corresponding editing intensity coefficients.

Our main contributions are summarized as follows:
\begin{itemize} [leftmargin=*]

    \item We study the impact of face editing on online \textit{ad} enhancement, which edits the facial features in an \textit{ad} cover image by changing its latent face representations. We use a pre-trained StyleGAN2-FFHQ~\cite{stylegan2} model as well as its corresponding pre-trained GAN inversion model e4e~\cite{e4e} for face generation and embedding. 
    Specifically, coupled with facial landmark detection techniques, a face image detected from an \textit{ad} is first encoded into the latent space of the GAN generator through the GAN inversion encoder e4e. We then modify the face representation in the latent embedding space and feed it into the generator to obtain a semantically edited version, which is finally replacing the original face to generate the enhanced \textit{ad}.
    \item We collect the QQ-AD dataset which contains 20,527 online ads with visual and textual information as well as their click rates information, based on which we train a new click rates prediction model based on six types of features, including Style-GAN-based facial latent vectors, in addition to image and text embeddings. This is the first online ad CTR predictor that takes into account latent embeddings from GAN. Offline tests have verified the superiority of our predictor to a range of baselines only using image embeddings or using NIMA image quality assessment \cite{talebi2018nima}, implying the important connections of facial characteristics to \textit{ad} popularity. We open source the implementation of AdSEE\footnote{Code available at \url{https://github.com/LiyaoJiang1998/adsee}.}.

    \item 
    We use the SeFa~\cite{sefa} model to find $q$ semantic editing directions in the latent space of the GAN generator through eigenvalue decomposition of the weight matrix of the generator. Each selected direction corresponds to a semantic facial characteristic, e.g., smile, age, etc. Then, we use a genetic algorithm to search for the best editing intensities for all the identified directions. With the identified directions and their corresponding optimal intensities, we adjust an \textit{ad} to the best appearance that may lead to higher click rates.

\end{itemize}

We further perform extensive analysis to offer insights on what directions and intensities of semantic edits may improve \textit{ad} click rates. 
We found that a face oriented slightly downward, a smiling face, and a face with feminine features are more attractive to clicks according to the analysis.
 
AdSEE was integrated into the Venus distributed processing platform at Tencent and deployed for an online A/B test in the recommendation tab of the QQ Browser mobile app (a major browser app by Tencent for smartphones and tablets). We report the test results of AdSEE in the traffic of QQ Browser mobile app for a period of 5 days in 2022.
As click rate is an important metric to gauge user satisfaction and efficiency of the business, with human-aided ethics control and censoring, the online A/B testing results show that AdSEE improved the average click rate of general ads in the QQ Browser recommendation tab, verifying the existence of the relationship between
image style editing and ad popularity.

\section{Related Work}
\label{sec:related}
\textbf{Click-Through Rate Prediction.} 
A CTR predictor aims to predict the probability that a user clicks an \textit{ad} given certain contexts which play an important role in improving user experience %
for many online services, e.g., e-commerce sites, social media platforms, and search engines. Recent studies extract visual features from the cover image of \textit{ad} for better CTR predicting~\cite{chen2016deep,cscnn,liu2017deepstyle,yang2019learning,zhao2019you}. 
~\citeauthor{chen2016deep}~\cite{chen2016deep} apply deep neural network (DNN) on \textit{ad} image for CTR prediction. 
\citeauthor{cscnn} propose the CSCNN~\cite{cscnn} model to encode \textit{ad} image and its category information, and predict the personalized CTR with user embeddings. \citeauthor{li2020adversarial}~\cite{li2020adversarial} utilize multimodal features including categorical features, image embeddings, and text embeddings to predict the CTR of E-commerce products. 
The sparsity and dimensionality of features vary drastically among different modalities.
Therefore, it is crucial to effectively model the interactions among the features from different modalities~\cite{ctr_survey}.
AutoInt is shown to achieve great performance improvement on the prediction tasks on multiple real-world datasets. Thus, in this paper, we build a click rate predictor to estimate the averaged click rate of an \textit{ad} among advertising audience based on the best-performing AutoInt~\cite{autoint} model compared with many state-of-the-art models in Appendix Section~\ref{sec:rec_models_compare}.

\textbf{Creatives Selection}. 
Another research direction of display advertising focuses on creatives selection. Previous studies in this line of research use the bandit algorithm for the news recommendations~\cite{li2010contextual}, page optimization~\cite{wang2016beyond}, and real online advertising~\cite{hill2017efficient}. \citeauthor{chen2021automated}~\cite{chen2021automated} propose an automated creative optimization framework to search for the optimal creatives from a pool. 
In this work, instead of choosing from various creatives, we enhance an existing \textit{ad} through direct facial feature editing. 

\textbf{Face Image Generation and Editing}. Generative Adversarial Networks (GANs)~\cite{goodfellow2014generative} have achieved impressive results on a range of image generation tasks. 
Style transfer~\cite{gatys2016image} is the task of rendering the content of one image in the style of another. 
StyleGAN~\cite{stylegan} proposes a style-based generator using the AdaIN~\cite{huang2017arbitrary} operation and can generate higher quality photo-realistic images compare to other alternatives~\cite{brock2018large,karras2017progressive}. 
Based on the StyleGAN2~\cite{stylegan2} model, \citeauthor{e4e} propose the e4e~\cite{e4e} encoder to map real face images to the latent embedding space of the StyleGAN2-FFHQ~\cite{stylegan2} model. \citeauthor{sefa} propose a closed-form factorization method to find the latent directions of face image editing without supervision. 
Following the style transfer~\cite{gatys2016image} direction, \citeauthor{durall2021facialgan} propose FacialGAN~\cite{durall2021facialgan} to transfer the style of a reference face image to the target face image. 
However, FacialGAN requires a standard face image as reference, which can not be satisfied when we have arbitrary faces in $ad$ cover images. Instead, our work utilizes the SeFa~\cite{sefa} image editing method to find the face editing directions without any supervision or reference images which is automated and efficient. To adjust the $ads$ to their best appearances that may lead to higher click rates, we find the optimal face editing intensity through the guidance of the predicted click rate. We adopt StyleGAN2 as our backbone image generation model because StyleGAN2 offers state-of-the-art generation quality and is applicable to many domains including faces, cars, animals, etc. Many works have chosen to extend StyleGAN2 including \cite{sefa, e4e, ganspace, psp} thus allowing many possible applications including image editing with SeFa~\cite{sefa}.

\section{Method}
\label{sec:method}

\begin{figure*}[ht]
  \centering
  \includegraphics[width=6.0in]{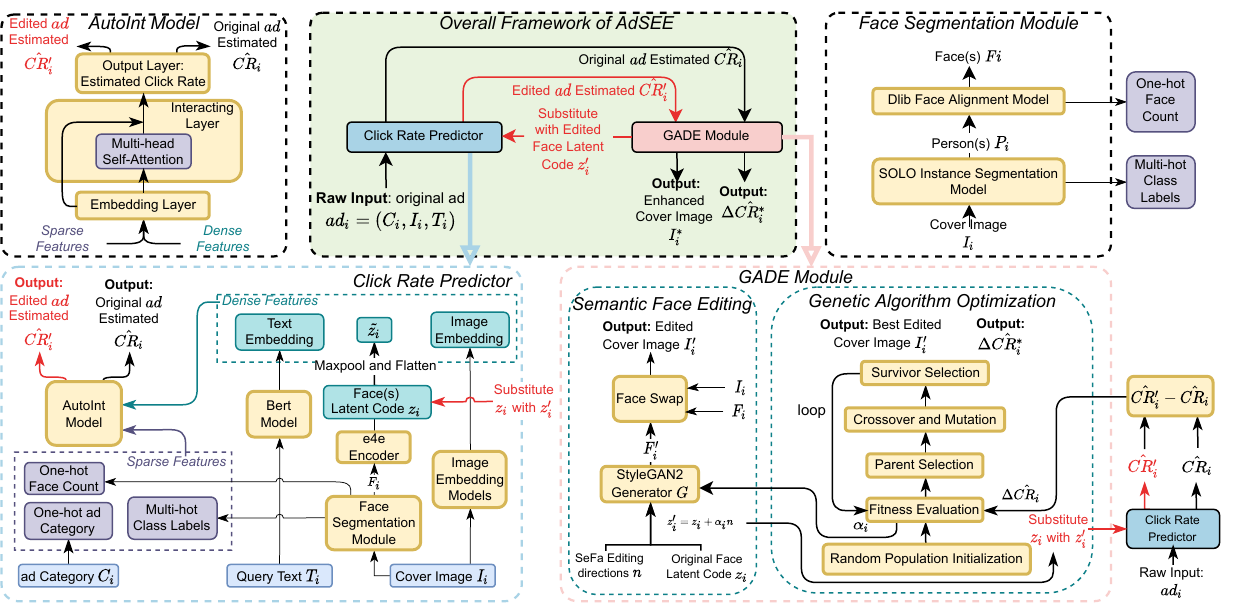}
  \caption{
  The system architecture of the proposed framework.
  }
  \Description{A figure shows the overall structure of the Click Rate Predictor.}
  \label{fig:sys}
\end{figure*}

In this section, we describe the detailed model adopted in the AdSEE framework.

We consider advertisement (\textit{ad}) data with category information, cover image, and query text. \blue{Each advertisement is displayed to the user within the app feed as a card which includes a cover image and a query text as the advertisement title.} Specifically, for a given advertisement $ad_i = (C_i, I_i, T_i), C_i \in \mathcal{C}$ where $C_i$ represents the category, e.g., ``Sports'', ``Game'', that $ad_i$ belongs to, $\mathcal{C}$ denotes the set of all the considered categories, and $I_i, T_i$ represent the cover image and query text of $ad_i$, respectively. 
\red{An \textit{impression} refers to the event when an ad is shown/exposed to a target user by the online advertising system.}
Therefore, to assess the attractiveness of $ad_i$, we calculate its averaged click rate as 
\begin{equation} \label{eq:ct_avg}
CR_{i} = \frac{click_i}{impression_i},
\end{equation}
where $click_i$ and $impression_i$ denote the total numbers of clicks and impressions of $ad_i$, respectively. 
The averaged click rate $CR_{i}$ indicates the overall attractiveness of $ad_i$ among the \textit{ad} audience. 

\subsection{System Overview}
Figure~\ref{fig:sys} provides an overview of our proposed AdSEE framework. 
First, we build a Click Rate Predictor (CRP) which takes an \textit{ad} as input and predicts its averaged click rate defined in~\eqref{eq:ct_avg}. Trained with a regression task, the CRP estimates the click rate of any given $ad$ which can be used to guide the ad editing module. 
Second, we build the Genetic Advertisement Editor (GADE) module to enhance the overall attractiveness indicated by $CR_i$ of $ad_i$ through editing its cover image $I_i$. The GADE module utilizes genetic algorithm to explore human facial feature editing directions in the form the face latent codes. It aims to find the best editing direction and editing intensities which may lead to the highest attractiveness enhancement reflected by the increase in predicted Click Rate with guidance from the CRP.

\subsection{Click Rate Predictor}
\label{subsec:method_ctr}
As shown in Figure~\ref{fig:sys}, we extract sparse and dense features from the raw input \textit{ad} data, i.e., $(C_i, I_i, T_i)$ and use the AutoInt~\cite{autoint} model structure to predict the average click rate for $ad_i$. 

\textbf{Sparse Features}. The category information of an \textit{ad}, e.g., ``Game'' is encoded as a one-hot vector, e.g. ``[0,1,0]''. The length of the encoded vector depends on the size of the category set, i.e., $|\mathcal{C}|$. 

The content of the cover image is also crucial to its overall attractiveness. Therefore, apart from the \textit{ad} category, we further extract sparse features from the cover image of an \textit{ad}. 
Specifically, we adopt the SOLO instance segmentation model \cite{solo,solov2} to identify the segmentation masks of all instances which belong to the COCO~\cite{lin2014microsoft} class, e.g., person, cat, etc. Formally, for an advertisement $ad_i = (C_i, I_i, T_i)$, we have
\begin{equation} \label{eq:SOLO}
\begin{aligned} 
 \textrm{Instance}_i &= \textrm{SOLO}(I_i) \\
 \textrm{Class}_i &= \textrm{Unique}(\textrm{Instance}_i),
\end{aligned}
\end{equation} 
where $\textrm{Instance}_i$ is the list of all detected instances by the SOLO model from the cover image $I_i$, and $\textrm{Unique}(\textrm{Instance}_i)$ identifies all the unique COCO classes, $\textrm{Class}_i$, from the instance list. The SOLO model supports the detection of 80 classes of COCO object labels. Therefore, we convert the detected $\textrm{Class}_i$ to a multi-hot encoded vector of size 80, e.g., $[0,1,0,...,0,1]$, where each 1 indicates the presence of a certain COCO class in the cover image. 

For instances that fall in the ``Person'' COCO class, we extract their corresponding person images according to their segmentation masks. Specifically, for a person instance, we apply Gaussian Blur\cite{gedraite2011investigation} to the unmasked area (non-person area) to blur the background out and isolate individual person to obtain person image, i.e., $P_{i,j}$ for the j-th person in the cover image $I_i$. 
Then, we feed all the person images, i.e., $ P_i = \{P_{i,j}\}, j = 1, \cdots, K_i, $ where $K_i$ is the total number of persons in cover image $I_i$, into the Dlib~\cite{dlib} face alignment model to align the facial landmarks and crop to face which yields face images $F_{i} = \{ F_{i,j}\}, j = 1, \cdots, M_i $, where $M_i \le K_i$ represents the number of detected faces from the $K_i$ person images $P_{i}$. That is, %
\begin{equation} \label{eq:Dlib}
\begin{aligned} 
 F_i &= \textrm{Dlib}(P_i).
\end{aligned}
\end{equation} 
Note that, we remove $ads$ that do not contain a face image because we cannot perform facial feature editing if there is no face in a cover image. 
In addition, we remove \textit{ads} with more than $M=5$ faces from the dataset to avoid extracting low-resolution and unrecognizable face images from a cover image. 
Thereafter, we encode the face count, i.e., $M_i$ where $ 1 \le M_i \le 5$, into a one-hot sparse vector with the length of 5, for example, a face count vector $[0,1,0,0,0]$ indicates 2 faces are detected from a cover image.

\textbf{Dense Features}. 
We further extract dense features from the cover image and query text of an \textit{ad} for the click rate prediction. 

First, we adopt the e4e model \cite{e4e} to encode each face image, $F_{i,j}$, into a real-valued dense vector representation $z_{i,j}$. Formally, for the j-th face image of cover image $I_i$, we have
\begin{equation} \label{eq:face_embedding}
z_{i,j} = E(F_{i,j}), j = 1, \cdots , M_i,
\end{equation}
where $E$ denotes the pre-trained e4e encoder for GAN inversion~\cite{xia2021gan} to the StyleGAN2-FFHQ \cite{stylegan2} face latent space, and $z_{i,j} \in \mathcal{R}^{d \times l}$ is the corresponding two-dimensional latent representation of face $F_{ij}$, and $M_i$ is the number of detected faces in cover image $I_i$. Then, we stack the $M_i$ latent embeddings of shape $[d, l]$ into one tensor of shape $[M_i, d, l]$, i.e.,
\begin{equation} \label{eq:stack}
z_{i} = [z_{i,1}, \cdots, z_{i,M_i}]. 
\end{equation}
We apply the best-performing max-pooling operation (among max-pooling, average-pooling, and concatenation operations) on $z_{i}$ along its first dimension to obtain the latent face code $\bar{z}_i$ with the shape of $[d,l]$. 
Then, $\bar{z}_i$ is flattened and used as a dense feature for the click rate prediction. 
With the latent representation $\bar{z}_i$, the CRP encodes the attractiveness of \textit{ads} from their facial features which enables using it to guide the GADE module for face style editing and cover image enhancement of \textit{ad}.

Second, apart from facial features, the attractiveness of an \textit{ad} also relies on the overall content and quality of the cover image. Therefore, we encode the whole cover image into a latent image representation to boost the click rate prediction. Specifically, we use two different image embedding methods to get more comprehensive and effective embeddings of the cover images. 
1) We adopt the image embedding model which is pre-trained with the multi-label image classification task on the open image dataset~\cite{OpenImages}. With more than 9.7 million images and around 20 thousand labels, the embedding provided by the multi-label classifier carries fine-grained image content information.
2) Another method of cover image embedding is provided by Sogou\footnote{A technology subsidiary of Tencent that provides search services.}. Sogou provides the service of searching pictures through text, in which both text and pictures are encoded into latent vectors for picture and text matching. Therefore, the embedding of the cover image provided by Sogou contains semantic information which is useful in judging the attractiveness of the cover image.
For a given \textit{ad}, we concatenate the image embeddings from the above two models to obtain the final embedding of the cover image.

Finally, we use a pre-trained Bert-Chinese model~\cite{bert} to extract text embedding from the query text, $T_i$, associated with $ad_i$. The Bert model takes the query text as input and outputs the embeddings of the words in the text. Then, we apply the max-pooling operation on the word embeddings to get the embedding of the query text. 

\textbf{Click Rate Prediction}
Let $x_{i,1}, \cdots, x_{i,6}$ denotes the 6 extracted features, including sparse features, \textit{ad} category, multi-hot class label, one-hot face count, and dense features, latent face representation, cover image embedding, and text embedding, for $ad_i$. 
Then, We apply the AutoInt~\cite{autoint} model to the extracted features, $x_{i,1}, \cdots, x_{i,6}$, to predict the averaged click rate. \blue{We selected the best-performing AutoInt model in our evaluation of many SOTA models in Appendix Section~\ref{sec:rec_models_compare}}.

For a given advertisement $ad_i$, to allow the interaction between sparse and dense features, the Embedding layer of the AutoInt model maps all 6 extracted features into a fix-length and low-dimensional space through embedding matrices, i.e., 
\begin{equation} \label{eq:embedding_layer}
\bm{h}_{i,k} = \textrm{Embed}(x_{i,k}), k=1, \cdots, 6, 
\end{equation}
where $\bm{h}_{i,k}$ denotes the low-dimensional feature of $x_{i,k}$, and $\textrm{Embed}(\cdot)$ is the standard embedding layer seen in almost all recommenders which learns a set of Embedding Weight Matrices, one for each input feature. Then, self-attention layers are adopted to model high-order feature interactions in an explicit fashion:
\begin{equation} \label{eq:attention}
\hat{\bm{h}}_{i,1}, \cdots , \hat{\bm{h}}_{i,6} = \textrm{Attention}(\bm{h}_{i,1}, \cdots , \bm{h}_{i,6}), 
\end{equation}
where $\textrm{Attention}(\cdot)$ denotes multiple self-attention layers, and $\hat{\bm{h}}_{i1}$ represents the high-order interaction features. 
Finally, the first-order features and their high-order interactions are fed into an output layer for click rate prediction. 
\begin{equation} \label{eq:output_layer}
\hat{CR}_i = \textrm{FC}(\hat{\bm{h}}_{i,1}^{\frown} \hat{\bm{h}}_{i,2}^{\frown} \cdots ^{\frown} \hat{\bm{h}}_{i,6} \oplus \bm{h}_{i,1}^{\frown}\bm{h}_{i,2}^{\frown} \cdots ^{\frown} \bm{h}_{i,6}),
\end{equation}
where $\bm{h}_{i,1}^{\frown}\bm{h}_{i,2}^{\frown}\cdots ^{\frown} \bm{h}_{i,6}$ denotes the vector after concatenating the vectors $\bm{h}_{i,1}, \cdots, \bm{h}_{i,6}$, and $\oplus$ represent point-wise addition. 
The output, $\hat{CR}_i$, of the fully-connected layer, $\textrm{FC}(\cdot)$, denotes the predicted average click rate of $ad_i$. 

The loss function of the Click Rate Predictor (CRP) is defined as the mean square error (MSE) between the predicted click rate and the target click rate: 
\begin{equation} \label{eq:loss_mse}
\mathcal{L} = \frac{1}{N}\sum_{i = 1} ^{N} ||\hat{CR}_i - CR_i||_2 + \alpha ||W||_2, 
\end{equation}
where $N$ is the number of \textit{ads}, and $W$ denotes the model weights. The first term represents the averaged square error between predicted click rates and the target click rates on the whole dataset. The second term is a regularizer to prevent over-fitting. The $\alpha$ is a hyperparameter that controls the influence of regularization. 

\subsection{Genetic Advertisement Editor}
\label{subsec:method_editing}

As shown in Figure \ref{fig:sys}, the Genetic Advertisement Editor (GADE) module takes the original faces latent code $z$ as input, iterates over generations of Face Image Enhancement guided by the CRP, then outputs the best-edited cover image $I_i^*$ and the best change in predicted click rate denoted as $\Delta\hat{CR_i^*}$. We describe the Semantic Face Editing module and the Genetic Algorithm Optimization (GAO) module in detail below.

\textbf{Semantic Face Editing.} Following \citeauthor{e4e}~\cite{e4e}, we adopt the closed-form semantic factorization method SeFa \cite{sefa} to identify a set of edit directions $n$ from the latent space of the pre-trained StyleGAN2-FFHQ \cite{stylegan2} face image generator $G(\cdot)$.
SeFa utilizes eigen-decomposition on the matrix $A^TA$, where $A$ is the weight matrix of $G(\cdot)$, to find a set of edit directions, i.e., $n=\{n_p\}^{q}_{p=1}$ where $n_p$ corresponds to the eigenvector associated with the $p$-th largest eigenvalue of the matrix ${A^T}A$. Each edit direction $n_p\in R^{512}$ corresponds to some face semantic concept, e.g. smile, eye-openness, age.

With the identified edit directions $n$, we apply the $edit(\cdot)$ operation to the face set $F_i$ to edit the facial image styles and enhance the attractiveness of a given cover image $I_i$. 
Formally, we have 
\begin{equation} \label{eq:edit}
    F_{i,j}' = edit(F_{i,j}) = G(z_{i,j}') = G(z_{i,j} + \alpha_{i,j} n), j = 1, \cdots, M_i, 
\end{equation}
where we alter the face image $F_{i,j}$ by linearly moving its original face latent codes $z_{i,j}$ along the identified direction $\alpha_{i,j} n$. Then, we use $G(\cdot)$ to generate edited face images $F'_{i,j}$ from edited face style vectors $z'_{i,j}$. 
In addition, $\alpha_{i,j}\in\mathcal{R}^{q}$ is the editing intensity coefficients given by a genetic algorithm for face image $F_{i,j}$, and $\alpha_{i,j} n$ denotes the linear combination of the $q$ edit directions $n$.

\textbf{Cover Image Editing}.
Then, we use the OpenCV \cite{opencv} and Dlib \cite{dlib} libraries to swap the edited face images $F_i'$ back into $I_i$ to obtain the edited cover image $I_i'$. 
The face swap operation $Swap(\cdot)$ can be formulated as
 \begin{equation} \label{eq:swap}
    I_i'=Swap(F_i,F_i',I_i),
\end{equation}
where the edited face images $F_i' = \{F_{i,j}\}_{j=1}^{M_i}$ are defined in~\eqref{eq:edit}. 

We measure the attractiveness enhancement of the edited faces $F_i'$ over the original faces $F_i$ using the difference in the predicted click rates of $F_i$ and $F_i'$, i.e.,
\begin{equation} \label{eq:delta_cr}
\begin{aligned}
    \Delta\hat{CR_i}&=\hat{CR_i'}-\hat{CR_i} \\  
\end{aligned}
\end{equation}
where $\hat{CR_i'}$ is the predicted average click rate of the edited cover image $I_i'$, and $\hat{CR_i}$ is the predicted average click rate of the original cover image $I_i$ defined in~\eqref{eq:output_layer}. 
Therefore, the enhancement of the attractiveness depends on the editing intensity coefficients $\alpha_i=\{\alpha_{i,j}\}^{M_i}_{j=1}$ and the identified editing directions $n$.

\SetKwInput{KwInput}{Input}
\SetKwInput{KwParameters}{Parameters}
\SetKwInput{KwGenotype}{Genotype}
\SetKwInput{KwInitialization}{Initialization}
\SetKwInput{KwFitnessFunction}{Fitness Function}
\SetKwInput{KwGenerationLoop}{Generation Loop}
\SetKwInput{KwOutput}{Output}

\begin{algorithm}[tb]
\DontPrintSemicolon
\caption{Genetic Advertisement Editor (GADE)}
\label{alg:algorithm}

\KwInput{Given $ad_i=(C_i,I_i,T_i)$; \\  
Set of original face latent codes $z_i = \{z_{i,j}\}_{j=1} ^{M_i}$; \\  
Set of SeFa Edit directions: $n=\{n_p\}^{q}_{p=1}$.}

\KwParameters{${PopulationSize}$, ${NumGeneration}$, \\ ${NumParents}$, ${PercentMutation}$.}

\KwGenotype{$\alpha_i=\{\alpha_{i,m}\}^{M_i}_{m=1}$, $\alpha_{i,m}\in\mathcal{R}^{q}$.}

\KwFitnessFunction{}
Fitness measurement $ \beta(\alpha_i)$ defined in~\eqref{eq:fitness}.

\KwInitialization{Generate the initial population $pop_1$ by \\
randomly generating $PopulationSize$ of genotypes.}

\KwGenerationLoop{}
\For{ $ \forall{g} \in [1, NumGeneration]$}
{
\textit{Fitness Evaluation:}
\green{evaluate the fitness for each genotype in $pop_g$ with $\beta(\cdot)$.}

\textit{Parent Selection:}
\green{use rank selection method to select $NumParents$ parents from $pop_g$ for mating.}

\textit{Crossover:}
\green{apply the uniform crossover operation among the parents to create off-springs.}

\textit{Mutation:}
\green{apply the random mutation operation to $PercentMutation$ percent of off-spring genotypes.}

\textit{Survivor Selection:}
\green{keep all $NumParents$ parents, and keep at most $PopulationSize-NumParents$ fit genotypes from the off-springs. All the kept genotypes are treated as the next population $pop_{g+1}$.}
}

\KwOutput{The best genotype $\alpha_i^*$.}

\end{algorithm}

\textbf{Genetic Algorithm}. 
To maximize the attractiveness enhancement $\Delta\hat{CR_i}$ defined in~\eqref{eq:delta_cr}, we adopt the genetic algorithm to search for the optimal editing intensity coefficients $\alpha_i^*$ for all the detected faces $F_i$ in cover image $I_i$. \blue{We selected the genetic algorithm to optimize the editing intensities because of its efficiency and effectiveness in a large search space. Alternatively, a gradient-based optimization approach will require backward passes through many components including the large generator model which is prohibitively expensive.}
Then, we generate the best-edited cover image $I_i'^*$ according to to~\eqref{eq:swap}.

We summarize the searching procedure of the genetic algorithm in Algorithm \ref{alg:algorithm}. 
We set the editing intensity coefficients $\alpha_i$ for $ad_i$ as the genotype in the genetic algorithm. 
Then, the fitness measurement $\beta_i(\alpha_i)$ for genotype $\alpha_i$ is set to be the predicted click rate $\hat{CR_i'}$ defined in~\eqref{eq:output_layer}. That is, 
\begin{equation} \label{eq:fitness}
\begin{aligned}
    \beta(\alpha_i)=\hat{CR_i'}=\textrm{Predictor}((C_i, I_i', T_i)) \\  
\end{aligned}
\end{equation}
where the $\textrm{Predictor}(\cdot)$ is the CRP, and $I_i'$, defined in~\eqref{eq:swap}, is the edited cover image of \textit{ad} $I_i$. 
Thus, guided with $\beta_i(\cdot)$, the genetic algorithm is supposed to search for the best genotype, i.e., editing intensity coefficients. 

At the initialization step, we create an initial population denoted as $pop_1$ by randomly generating PopulationSize number of genotypes, each with the same shape as $\alpha_i$.
Then, we repeat the generation loop $NumGereration$ times. Each iteration consists of five steps including \textit{Fitness Evaluation, Parent Selection, Crossover, Mutation,} and \textit{Survivor Selection}.

After $NumGereration$ generations, we return the best genotype $\alpha_i^*$ with the highest fitness value, 
which also results in the best improvement of the predicted click rate $\Delta\hat{CR_i^*}$ defined in~\eqref{eq:delta_cr}.  
Finally, we use the best genotype $\alpha_i^*$ to generate the best cover image $I_i$ according to~\eqref{eq:swap}. 

\section{EVALUATION}
\label{sec:results}

In this section, we conduct extensive experiments to evaluate the effectiveness of the proposed click rate predictor and the AdSEE framework. We also perform offline and online analyses based on the introduced QQ-AD dataset to offer insights on the connection between style editing and possible click rate enhancement. \blue{Furthermore, we also evaluate AdSEE on the public CreativeRanking~\cite{wang2021hybrid} dataset.} Due to space constraints, we qualitatively evaluate AdSEE edited images by putting examples in the Appendix Section~\ref{subsec:image_examples}.

\begin{table}[tbp]
\caption{Statistics of the Collected QQ-AD Dataset.}  
\begin{center}
\begin{tabular}{lcccc}
\toprule
Dataset    & \#Ads                     & \#Impressions     & \#Clicks & CR    \\
\hline
QQ-AD      & 158,829  & 4,263,667,016 & 429,830,278 & 0.1008\\
Applicable & 20,527   & 815,272,384 & 83,729,560 &  0.1027\\
\hline
Ratio            & 12.92\%  & 19.12\% & 19.48\% & -- \\
\bottomrule
\end{tabular}
\label{table:dataset_info}
\end{center}
\end{table}

\subsection{Datasets}

\textbf{QQ-AD Dataset.}
To evaluate our proposed approach, we collected real advertisement data from the QQ Browser mobile app. \blue{Note that the common recommender model datasets such as Avazu~\cite{avazu} and Criteo~\cite{criteo} do not apply to our work because they do not contain any image.} Each $ad$ record consists of its category information, cover image, and query text. 
In addition, we also collected the number of impressions, i.e., the number of times an \textit{ad} is shown to an audience, and the number of clicks, i.e, the number of times that an \textit{ad} was clicked by an audience. Shown in Table~\ref{table:dataset_info}, we collected a total number of 158,829 \textit{ads} from December 19, 2021, to January 18, 2022.
As our goal is to enhance the attractiveness of ad images through facial feature editing, we remove \textit{ads} that do not contain a face in their cover image. 
In addition, we also remove \textit{ads} with more than m=5 faces in its cover image from the collected dataset to avoid extracting low-resolution and unrecognizable face images from the cover image of an \textit{ad}.
Finally, we have 20,527 \textit{ads} with a valid number of faces in the collected QQ-AD dataset. That is, around 12.92\% of the collected \textit{ads} from the QQ Browser mobile environment contain 1-5 faces that can be enhanced with our AdSEE framework. 
The number of impressions and clicks for AdSEE applicable images in the QQ-AD dataset accounts for 19.12\% and 19.48\% of the total number of impressions and clicks, respectively. 
This suggests that an \textit{ad} image with 1 to 5 faces is common in the QQ Browser mobile environment, and editing the facial features can potentially have a significant impact on the overall user clicks, impressions, and click rates. We randomly split the \textit{ads} in QQ-AD dataset into three parts for training (64\%), validation (16\%), and testing (20\%).

\textbf{CreativeRanking Dataset.}
We further evaluate AdSEE on the relevant public dataset CreativeRanking\footnote{Dataset available at \url{https://tianchi.aliyun.com/dataset/93585}.} published by ~\citeauthor{wang2021hybrid} ~\cite{wang2021hybrid}. We process the CreativeRanking dataset to be similar to our image enhancement task. Each row in CreativeRanking dataset contains an e-commerce image, a product name, a number of clicks, a number of shows, and a show date. We aggregate the total clicks and the total shows for the same product and the same image over different dates resulting in each row corresponding to an image-product pair and the corresponding total show, total click, and average click rate. Similar to the features used in Section~\ref{sec:method}, we use the same one-hot face count (from 0 to 5) and multi-hot class label as the sparse feature, and we use face latent code and image embedding as the dense feature. Differently, we replace the $ad$ category sparse feature with the product name index as a sparse feature and we do not use any text embedding feature since there is no text data in CreativateRanking. In this dataset, there can be different images that are for the same product so each row in our dataset is a product-image pair. We remove any product-image pair with less than 100 total impressions, with more than 1000 total impressions, or with 0 total clicks. This yields 267,362 product-images pairs which we split into three parts for training (60$\%$), validation (20$\%$), and testing (20$\%$). We use the train set which contains both images with face and images with no face to train the CRP model. However, the GADE model should be applied to images containing faces, so we further filter any images with no faces or more than 5 faces resulting in a total of 23,713 valid images with a desirable number of faces in the entire dataset.

\subsection{Evaluation on QQ-AD and CreativeRanking}

In the offline evaluation, we first evaluate the proposed CRP model on the click rate prediction task and compare it against a wide range of baseline methods on both of the QQ-AD dataset and the CreativeRanking dataset. 
Then, we want to analyze whether style editing using the GADE module is linked to attractiveness and $ad$ popularity improvements. Thus, we edit the \textit{ads} with the GADE module and evaluate the improvement of the attractiveness, measured by $\Delta\hat{CR}$, of the edited \textit{ads} through the CRP model on both of the QQ-AD dataset and the CreativeRanking dataset.
Finally, we perform case studies to analyze the more attractive face editing directions on the QQ-AD dataset.

\begin{table*}[tbp]
\centering
\caption{Comparing the proposed CRP predictor with other baselines using different types of features on the QQ-AD dataset.}
\label{tab:crp-baseline}
\begin{tabular}{l|l|cccccc}
\hline
Model         & Feature Type     & MAE $\downarrow$   & MAPE $\downarrow$  & NDCG@10 $\uparrow$ & NDCG@50 $\uparrow$ & Spearman's rho $\uparrow$ & Kendall's tau $\uparrow$ \\ \hline
CRP-NIMA      & Image Quality    & 0.0299 & 0.7456 & 0.2764  & 0.3917  & 0.3634         & 0.2480        \\
CRP-OpenImage & Image Embedding  & 0.0295 & 0.7258 & 0.5950  & 0.5551  & 0.3941         & 0.2696        \\
CRP-Sogou    & Image Embedding  & 0.0299 & 0.7429 & 0.5095  & 0.5175  & 0.3613         & 0.2464        \\
CRP-e4e       & Face Latent Code & 0.0306 & 0.7663 & 0.5149  & 0.5204  & 0.2954         & 0.2003        \\ \hline
\textbf{CRP} & Combined & \textbf{0.0262} & \textbf{0.6542} & \textbf{0.6854} & \textbf{0.7337} & \textbf{0.5122} & \textbf{0.3609} \\ \hline
\end{tabular}
\end{table*}

\begin{table*}[tbp]
\centering
\caption{Comparing the proposed feature combination C5 with other combinations on the CreativeRanking~\cite{wang2021hybrid} dataset. We consider features including Face Count (FC), Product Name (PN), Class Label (CL), Face Latents (FL), and Image Embedding (IE).}
\label{tab:crp-baseline-cr}
\begin{tabular}{l|l|l|cccccc}
\hline
\# & Sparse Features & Dense Features & MAE $\downarrow$ & MAPE $\downarrow$ & NDCG@10 $\uparrow$ & NDCG@50 $\uparrow$ & Spearman's rho $\uparrow$ & Kendall's tau $\uparrow$ \\ \hline
C1 & FC, CL     & FL, IE         & 0.0134          & 0.5988 & 0.4567 & 0.4977 & 0.3374 & 0.2299 \\
C2 & PN, CL     & FL, IE         & \textbf{0.0132} & 0.6300 & 0.4975 & 0.4935 & 0.3304 & 0.2255 \\
C3 & FC, PN, CL & FL             & 0.0136          & 0.6479 & 0.3888 & 0.4073 & 0.2978 & 0.2020 \\
C4 & FC, PN, CL & IE             & 0.0135          & \textbf{0.5939} & 0.4865 & 0.4674 & 0.3379 & 0.2298 \\ \hline
\textbf{C5} & \textbf{FC, PN, CL} & \textbf{FL, IE} & \textbf{0.0132} & 0.5947 & \textbf{0.5065} & \textbf{0.5256} & \textbf{0.3609} & \textbf{0.2468} \\ \hline
\end{tabular}
\end{table*}

\begin{figure*}[hbtp]
    \centering
	\subfigure[Distribution of $\Delta\hat{CR}$ in the offline test on the QQ-AD dataset.]{
        \includegraphics[width=2.18in]{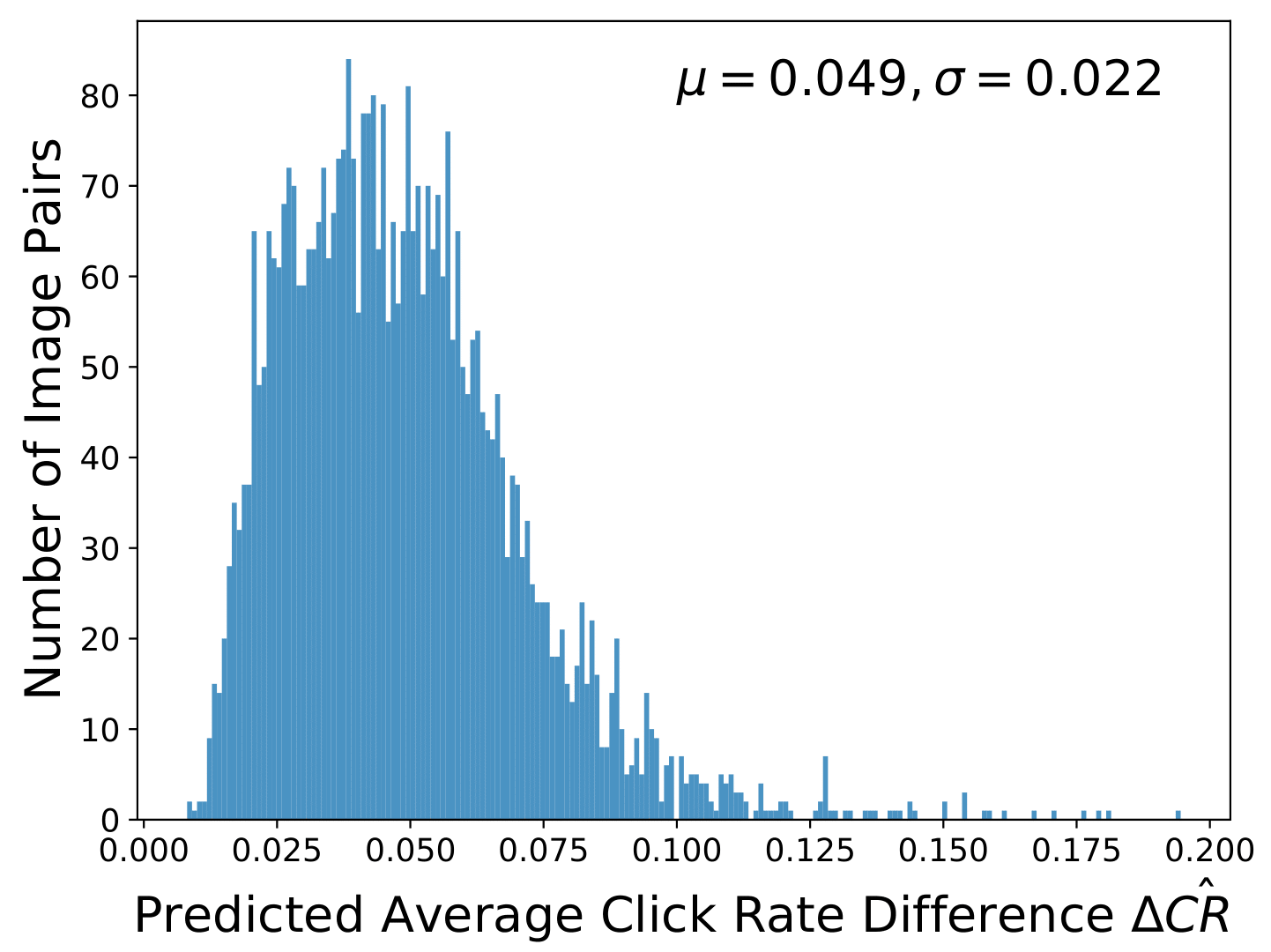}
        \label{fig:offline_compare_hist}
    }
    \hfill
    \subfigure[Distribution of $\Delta\hat{CR}$ for 10 different edit directions in the offline test on the QQ-AD dataset.]{
        \includegraphics[width=2.15in]{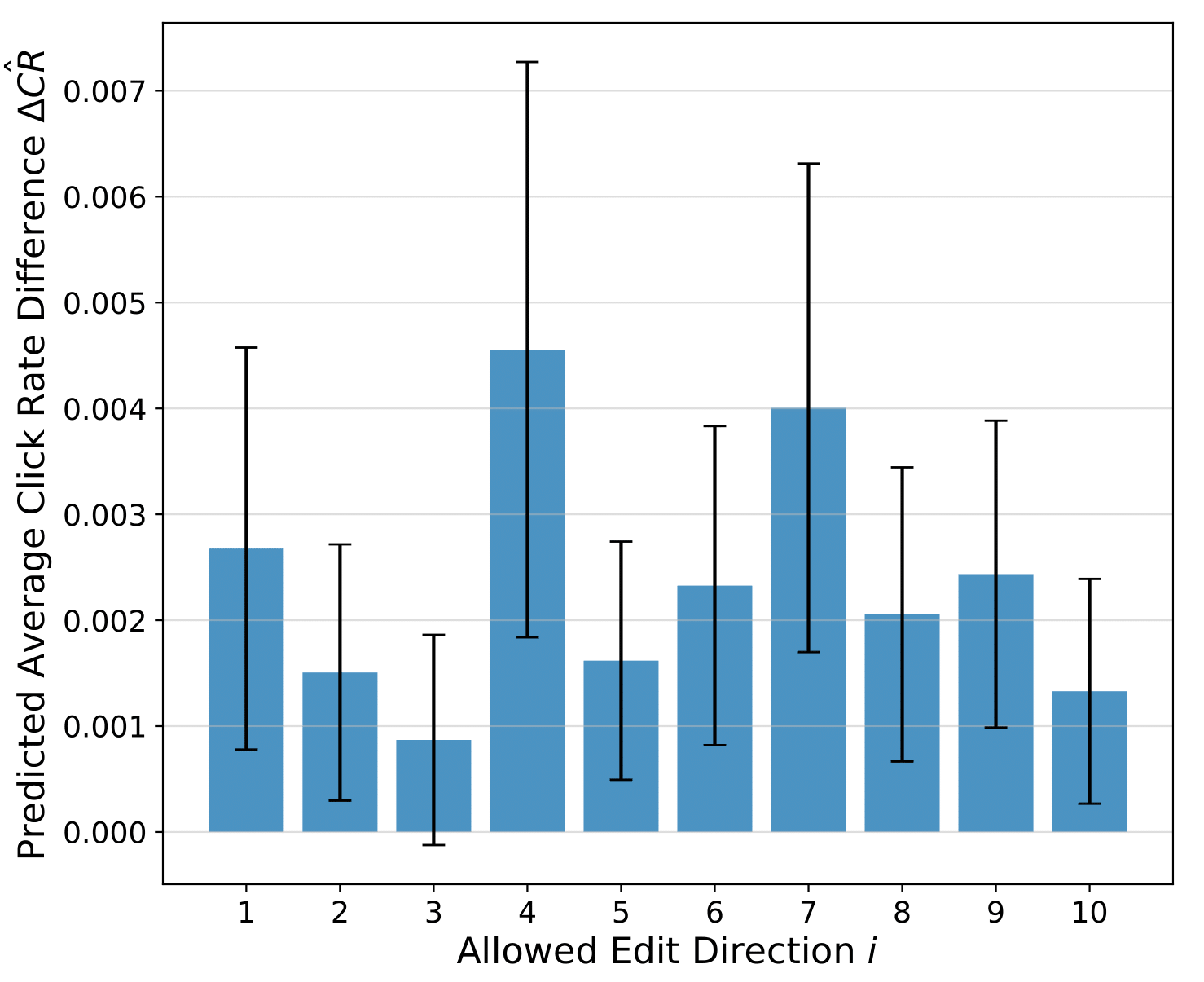}
        \label{fig:edit_compare_bar}
    }
    \hfill
    \subfigure[Distribution of $\Delta\hat{CR}$ in the offline test on the CreativeRanking~\cite{wang2021hybrid} dataset.]{
        \includegraphics[width=2.32in]{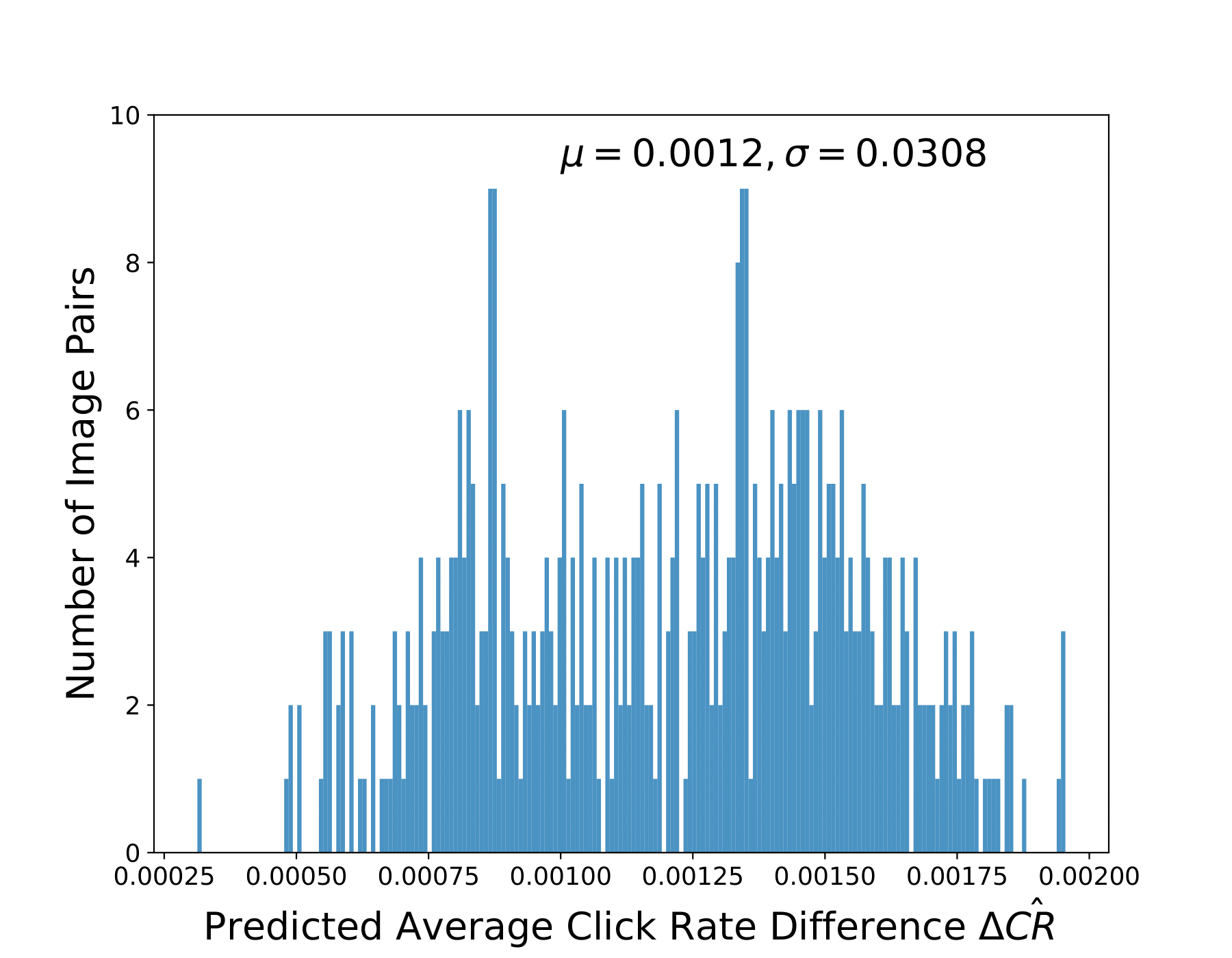}
        \label{fig:offline_compare_hist_cr}
    }
    \caption{Analysis of predicted average click rate difference $\Delta\hat{CR}$ in the offline evaluations.}
    \label{fig:offline_delta_cr_merged}
\end{figure*}

\textbf{Evaluation of the CRP model}
In this experiment, we compare the proposed CRP method with the following state-of-the-art baseline methods that use different features for average click rate prediction on the QQ-AD dataset.
\textbf{ CRP-NIMA:} This is the baseline method of \cite{talebi2018nima} where the NIMA score mean and standard-deviation are used as dense features.
\textbf{ CRP-OpenImage:} Use the image embeddings obtained from the multi-label image classification model pre-trained on Open Image dataset \cite{OpenImages} as dense features.
\textbf{ CRP-Sogou:} Use the image embeddings obtained from the Sogou model for searching pictures through text as dense features.
\textbf{ CRP-e4e:} Use the max-pooled face latent codes obtained from the pre-trained e4e FFHQ encoder \cite{e4e, stylegan} model as dense features.

The implementation details including hyperparameters, pre-trained models, and environment are introduced in Appendix Section~\ref{sec:reproducibility}. Note that, we train the proposed model and all the other baseline methods with the same MSE loss for a fair comparison. Moreover, the sparse features, i.e., \textit{ad} category, multi-hot class label, and one-hot face count, are used in all methods.
Furthermore, we adopt the mean absolute error (MAE), mean absolute percentage error (MAPE), normalized discounted cumulative gain (NDCG), Spearman's $\rho$, and Kendall's $\tau$ to evaluate the performance of different models for the average click rate prediction task. 

Table \ref{tab:crp-baseline} summarizes the performances of the proposed CRP model and all the baseline methods on the QQ-AD dataset. We can clearly see that our proposed CRP significantly outperforms all the other baselines on all the evaluation metrics. 
The superiority of the proposed method over other baselines can be attributed to the adoption of multi-modal dense features, i.e., face latent code, image embedding, and text embedding. 
Note that, the CRP-NIMA baseline method from ~\cite{talebi2018nima}, which uses image quality NIMA score as a dense feature, is the worst model in terms of NDCG@10 and NDCG@50 when compared against the rest of the methods where image embedding and face latent code is adopted as the dense feature. This, again, demonstrates the importance of our extracted dense features for the accurate click rate prediction of an \textit{ad} and the correlation between image style and ad popularity.

\blue{Table \ref{tab:crp-baseline-cr} summarizes the performances of the proposed CRP model and all the baseline methods on the CreativeRanking~\cite{wang2021hybrid} dataset. We train our CRP predictor on the preprocessed CreativeRanking dataset with MSE loss. We call the proposed combined set of features C5, which includes sparse features one-hot face count, one-hot product name, and multi-hot class label. In addition, C5 includes dense features face latent code, and image embedding. First, we compare the performance of different feature combinations on the CreativeRanking dataset and we found our proposed combined set of features (C5) outperforms all other feature combinations on 5 out of 6 metrics.  All instances use the AutoInt model and share the same training settings for a fair comparison. Results on both QQ-AD and CreativeRanking demonstrate the benefits of the multi-modal features we proposed to use. Specifically, using a combination of face latent vectors, image embeddings, and text embeddings can achieve better performance than the baseline features.}

\textbf{Evaluation of the GADE model}.
In this experiment, we want to answer the question: does editing facial styles of an $ad$ using our AdSEE model improve the attractiveness of an \textit{ad}?
An edited \textit{ad} and its corresponding original \textit{ad} will form an evaluation pair. 

In Figure~\ref{fig:offline_compare_hist}, we can observe that the values of the $\Delta\hat{CR}$ are positive for all the evaluation pairs in the test set of QQ-AD dataset. This shows that facial style editing do improve the attractiveness of an $ad$ through using our AdSEE framework. 
Furthermore, the $\Delta\hat{CR}$ has a mean of 0.049 and is right skewed which means that most of the samples have a relatively small positive increase in the predicted click rate, i.e., $\hat{CR}$, after being edited by the AdSEE model. Whereas, a few \textit{ads} have a large increase in the predicted click rate. This is reasonable because most of the cover images of the \textit{ads} are already well-designed and have decent attractiveness.

\blue{In addition, we use the GADE module together with the CRP predictor to optimize a random sample of 500 images from the CreativeRanking~\cite{wang2021hybrid} dataset test set (keeping images with 1 to 5 faces). We summarize the predicted average click rate difference $\Delta\hat{CR}$ in Figure~\ref{fig:offline_compare_hist_cr}. We observe the $\Delta\hat{CR}$ for all 500 test images are all positive and have a mean of $0.0012$ which is a 3.9\% increase relative to the 0.0308 mean $CR$ of these test images. This demonstrates that our method can enhance image attractiveness when applied to other image recommendation scenarios like e-commerce besides our own advertisement QQ-AD dataset. This shows AdSEE can be used to extract knowledge and can enhance image attractiveness by facial image style editing. Moreover, the results show the existence of a correlation between image style editing and click rates in $ads$.}

\textbf{Semantic Editing Directions.}
To figure out the most important semantic editing directions that improve the attractiveness of a cover image the most, we sample 1000 images from the QQ-AD test set and run AdSEE 10 times. 
In each run, we allow editing in only one out of the top ten directions discovered by SeFa \cite{sefa}. 
In Figure \ref{fig:edit_compare_bar}, we observe that editing on directions $n_4$, $n_7$, $n_1$ results in the largest increase in $\hat{CR}$. That is, these directions have the largest impact on the attractiveness of an \textit{ad} among the other editing directions. We further analyze the details on semantic editing directions in Figure \ref{fig:case_edit_direction} of the Appendix Section~\ref{subsec:edit_ditections}.

\subsection{Online A/B Tests}

\begin{figure*}[hbtp]
    \centering
	\subfigure[Click Rate per day (p-value $3.68 \times 10^{-5}$)]{
        \includegraphics[width=2.0in]{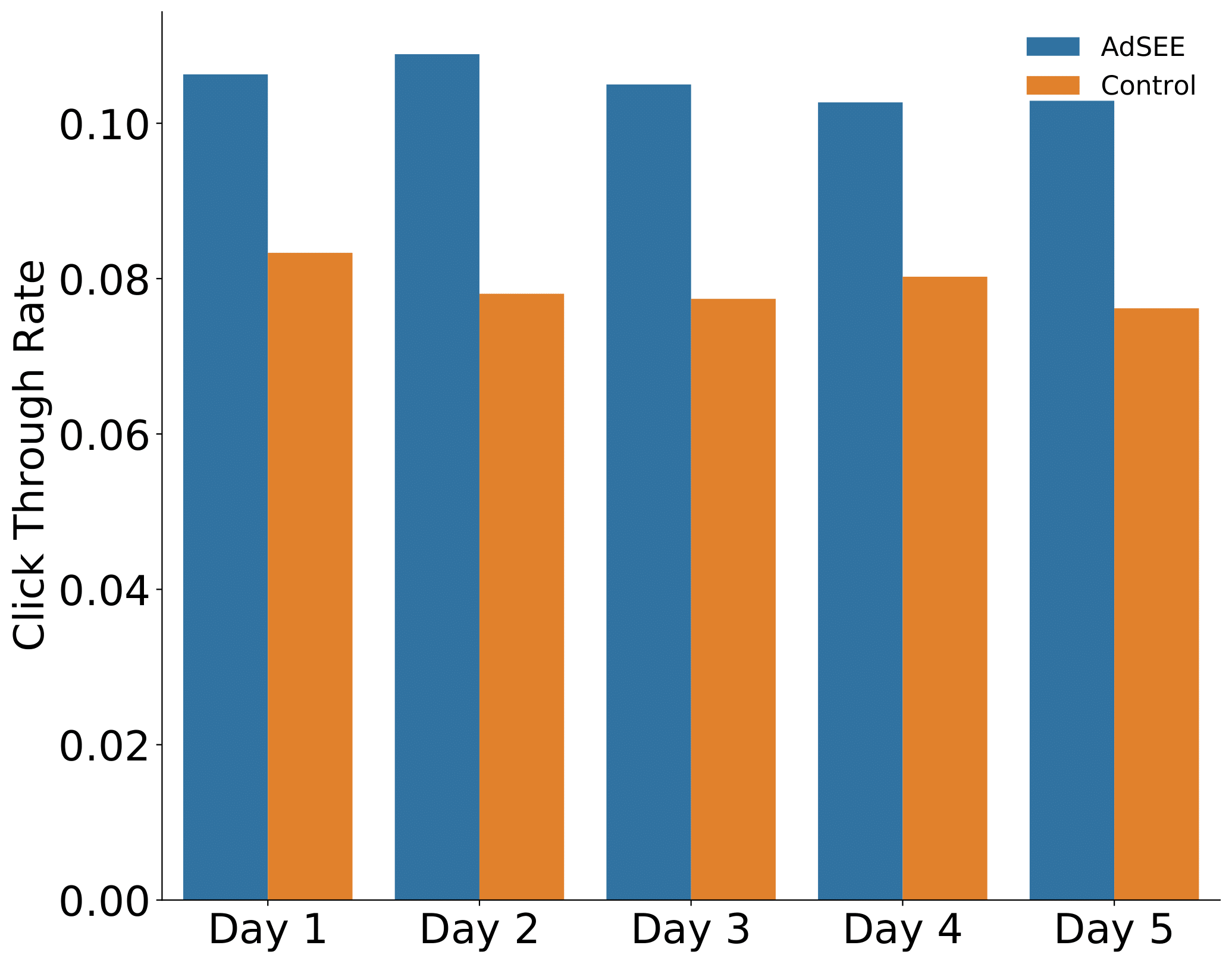}
        \label{fig:ctr_day}
    }
    \hfill
    \subfigure[Number of clicks per day (p-value $1.86 \times 10^{-2}$)]{
        \includegraphics[width=2.0in]{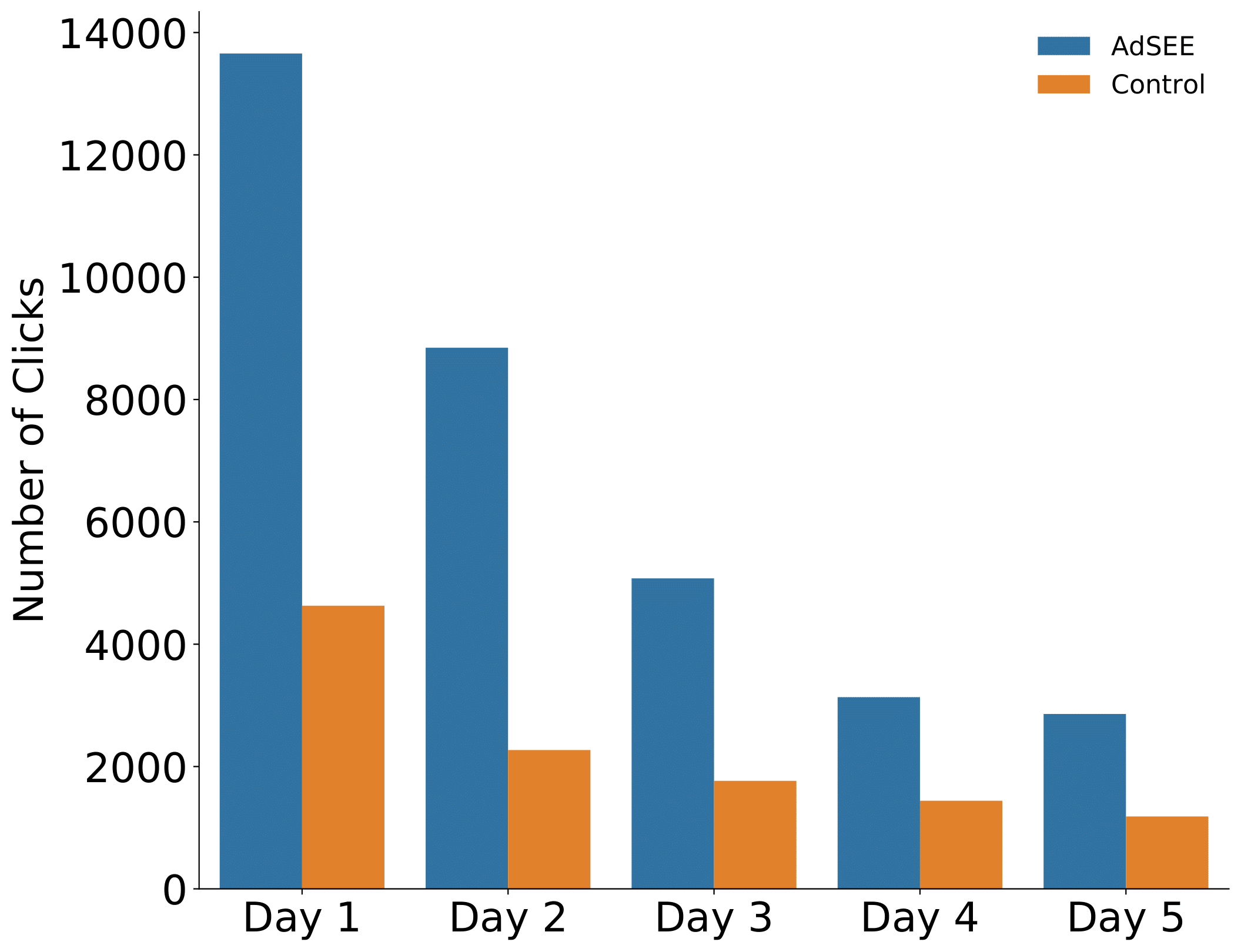}
        \label{fig:clicks_day}
    }
    \hfill
    \subfigure[Number of impressions per day (p-value $2.11 \times 10^{-2}$)]{
        \includegraphics[width=2.0in]{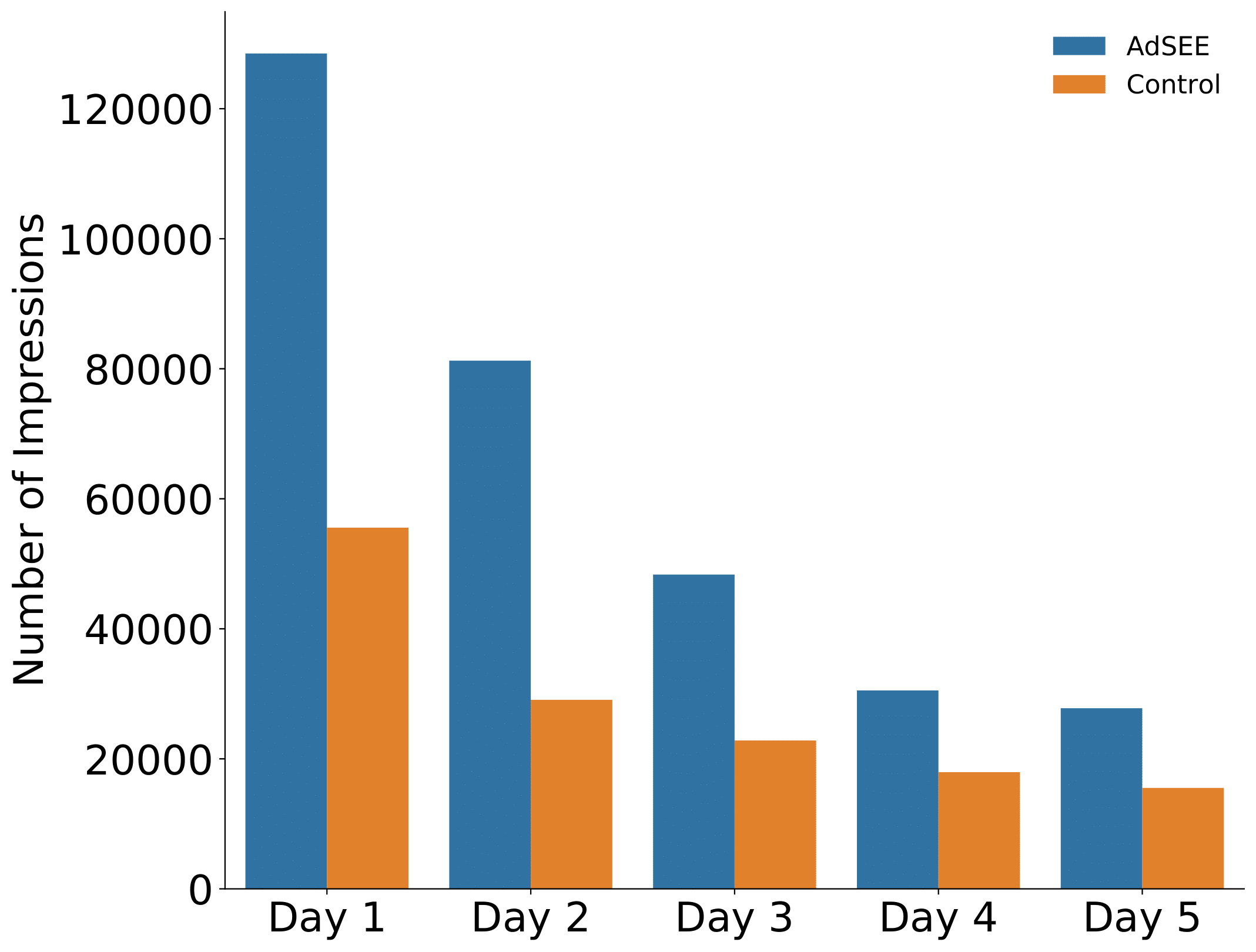}
        \label{fig:expose_day}
    }
    \caption{Comparison between AdSEE and control group in terms of the number of impressions, clicks and click rates.}
    \label{fig:day_plots}
\end{figure*}

\begin{figure*}[hbtp]
    \centering
	\subfigure[Click Rate Difference]{
        \includegraphics[width=2.0in]{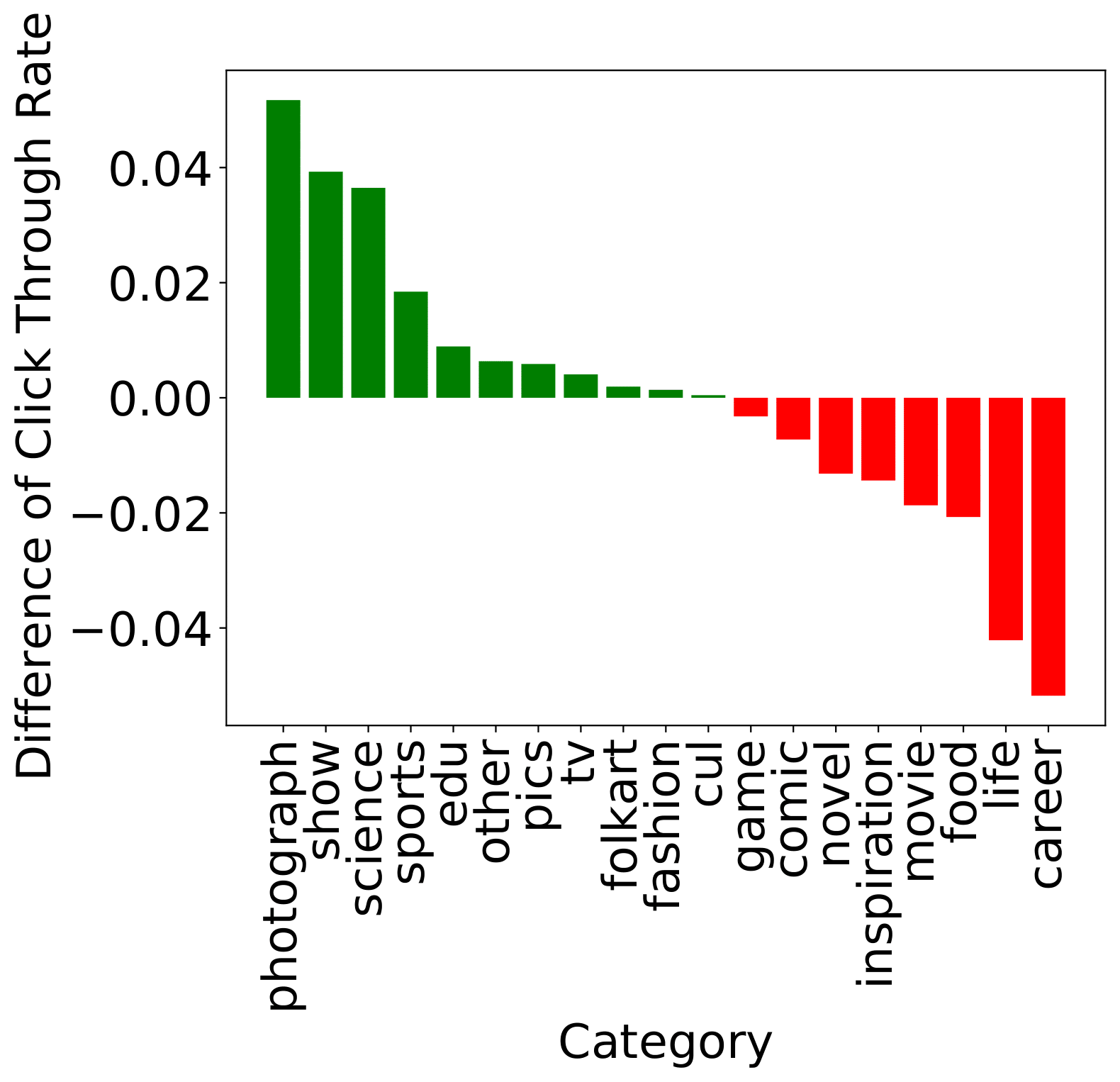}
        \label{fig:ctr_category}
    }
    \hfill
    \subfigure[The difference of number of impressions]{
        \includegraphics[width=2.0in]{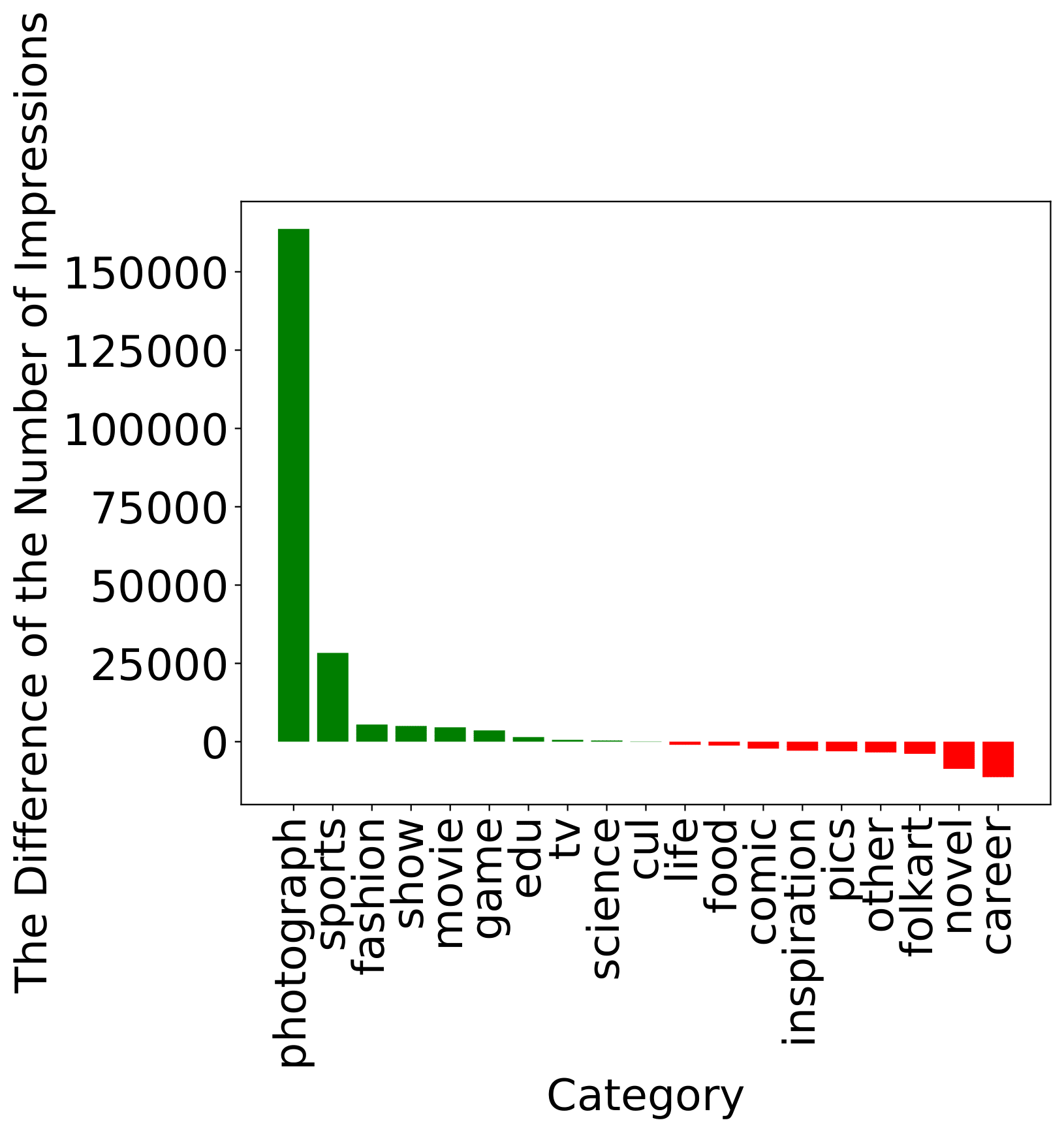}
        \label{fig:expose_category}
    }
    \hfill
    \subfigure[The difference of number of clicks]{
        \includegraphics[width=2.0in]{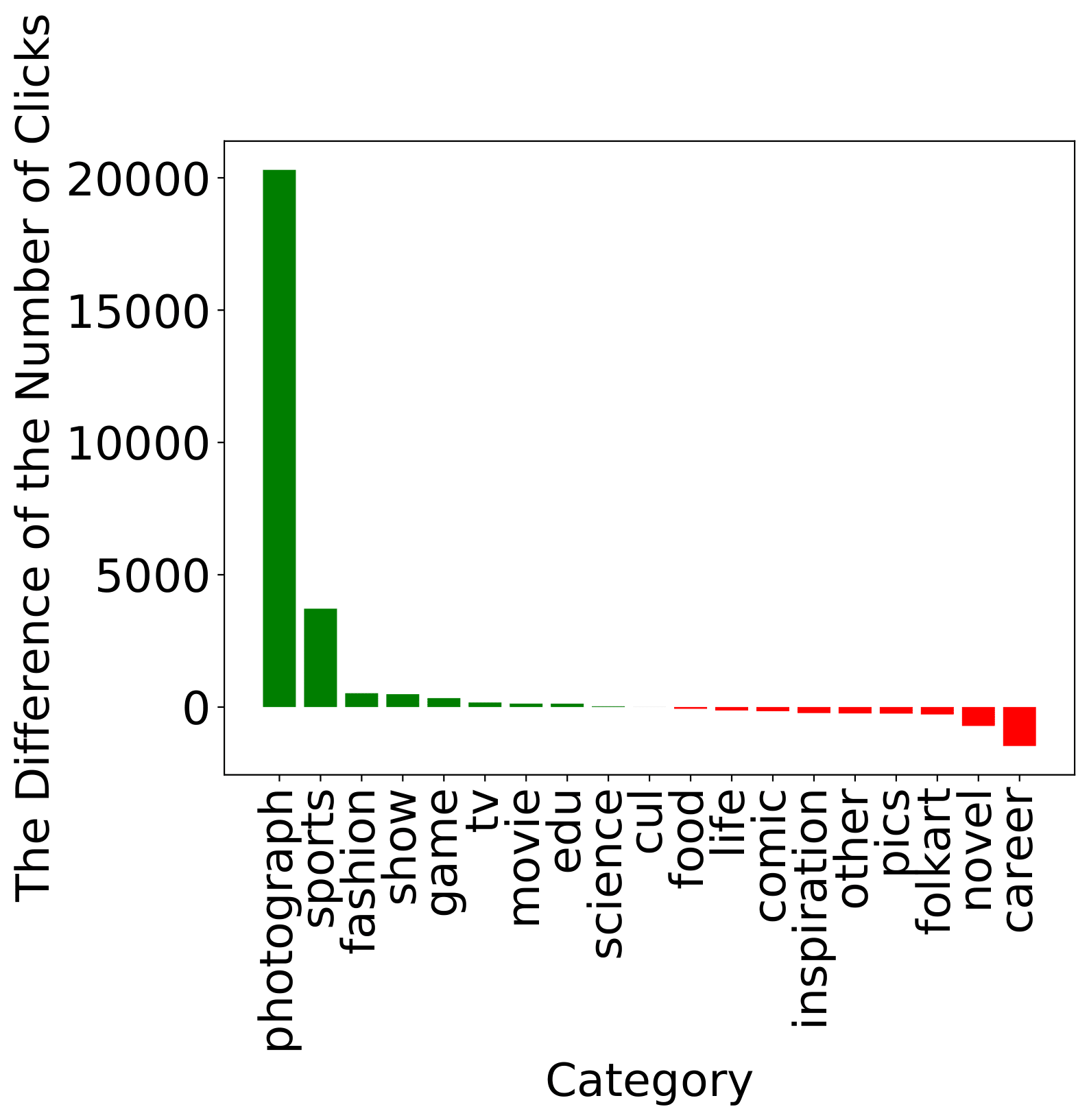}
        \label{fig:clicks_category}
    }
    \caption{The difference (increase) after applying AdSEE to each category of \textit{ads} in terms of the click rate, number of impressions, and number of clicks.}
    \label{fig:category_plots}
\end{figure*}

\begin{figure*}[hbtp]
    \centering
	\subfigure[CDF of the click rate]{
        \includegraphics[width=2.0in]{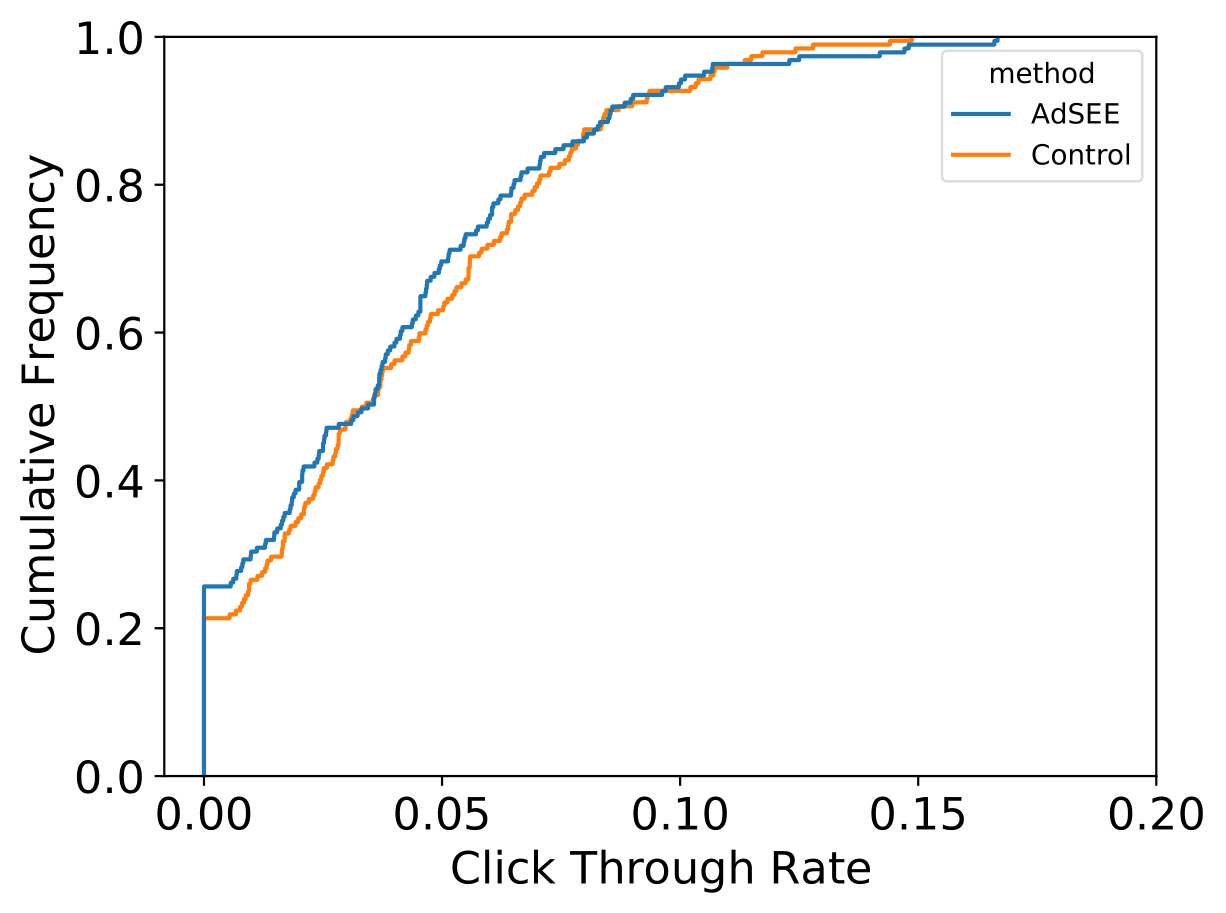}
        \label{fig:ctr_cdf}
    }
    \hfill
    \subfigure[CDF of the logarithm of the number of clicks]{
        \includegraphics[width=2.0in]{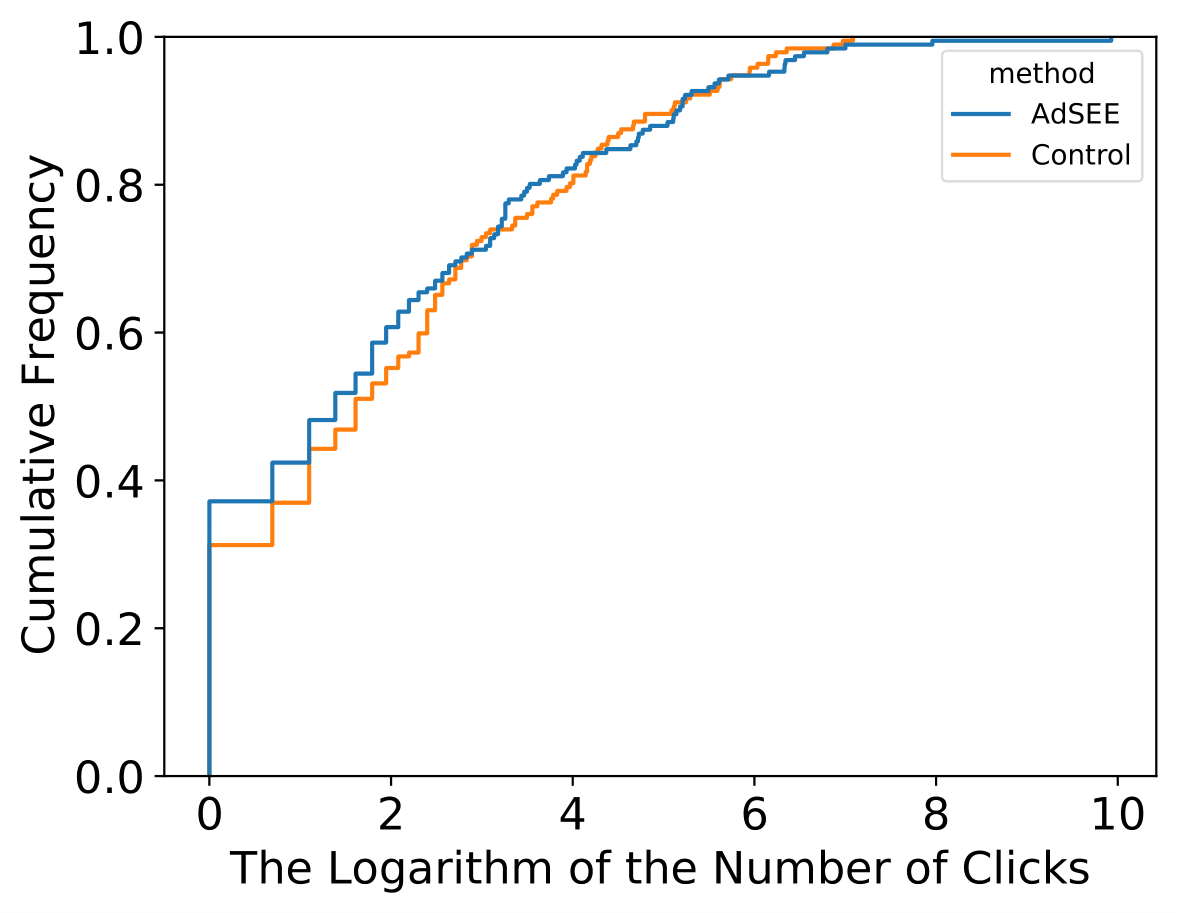}
        \label{fig:clicks_cdf}
    }
    \hfill
    \subfigure[CDF of the logarithm of the number of impressions]{
        \includegraphics[width=2.0in]{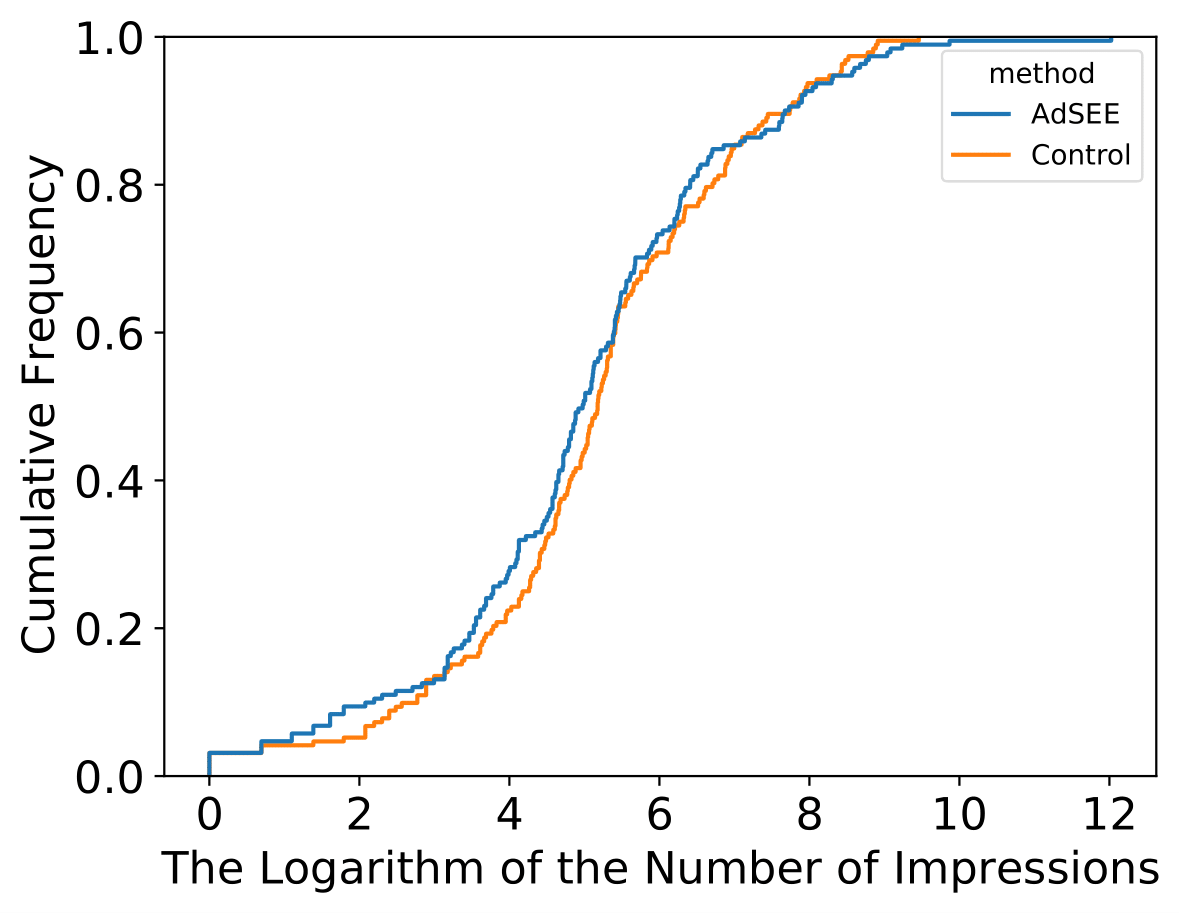}
        \label{fig:expose_cdf}
    }
    \caption{The cumulative frequency distribution of click rates, the logarithm of the number of impressions, and the logarithm of the number of clicks.}
    \label{fig:cdf_plots}
\end{figure*}

We further report the results of an online A/B test, by comparing 250 \textit{ad} images edited and altered by the AdSEE model as well as the 250 original \textit{ad} images tested over the QQ Browser mobile app users in a 5-day period. \blue{All of the 250 original images and the corresponding 250 edited images contain faces.} These $ads$ fall into 19 categories (genres), including photography, sports, fashion, show, game, TV, movie, education, science, culture, food, life, comic, inspiration, other, pics, folk arts, novel, and career. The online A/B tests were performed over a period of 5 days from Feb 5th, 2022 to Feb 9th, 2022, where we collected the number of impressions and clicks, and click  rates to compare AdSEE with the control group of unaltered images.
\red{Recall that an impression refers to
the event when an ad is shown/exposed to a user by the
online advertising system and that click rate equals to the number of clicks divided by the number of impressions.} 

The result shows that images edited by facial style editing with AdSEE received a significant increase in attractiveness compared to the original images in every metric. When performing online split tests, the AdSEE images and the original images were uploaded at the same time and presented on QQ Browser. After 5 days, we collected the number of impressions and clicks, and conducted the following comparisons. Figure~\ref{fig:day_plots} shows that AdSEE images were presented a larger number of times and showed a higher click rate hence attracting more clicks on ads each day separately. By conducting Paired Sample T-Test, we validated that the experiment group was significantly better than the control group. \blue{A larger number of impressions indicate that AdSEE-enhanced images are recommended more times by the recommender model, which means the independent production recommendation model that is not trained on AdSEE-edited images "believes" AdSEE-edited images are more attractive to users and may lead to increase in click rates.} In addition, a higher click rate indicates better attractiveness to users. Figure~\ref{fig:category_plots} shows the difference in performance in each of the 19 categories, we largely improved the popularity and attractiveness of images in the photograph category and sports category in terms of every metric. Figure~\ref{fig:cdf_plots} shows the cumulative frequency distribution of the number of impressions, click rate, and the number of clicks, the results indicate that AdSEE images outperformed the control group in different bins of images.

\section{Limitations and Discussion}
\label{sec:ethics}

This work represents one of the first efforts to explore the potential impact of art and image synthesis on recommender systems. Specifically, we aim to investigate if there is a linkage/correlation between popularity and image styles through a data-science approach, which we believe is a valuable question to ask for the AI community as well as AI ethics community. We verified the existence of this linkage with both offline experiments and online A/B testing. However, we do not aim to commercialize AdSEE as a traffic booster at the moment. In addition, any exploitation of the research results for commercial use is subject to further consideration of ethical requirements and regulations. 

A similar case also applies to recent advancements in content generation like image generation models StyleGAN3~\cite{stylegan3}, DALLE$\cdot$2~\cite{ramesh2022hierarchical}, and Imagic~\cite{kawar2022imagic}, which are widely popular in AI research because they can automatically generate state-of-the-art synthetic images that may match the quality of real images created by cameras and human artists.
\red{Meanwhile, we recognize that the nature of synthetic image generation tasks inherently brings risks to areas such as information objectivity, misleading information, copyright, data privacy, data fairness, etc. Therefore, we believe it is crucial that any research in the image generation area should be performed with broader societal and ethical impacts in mind.} 

We hold copyright protection, data privacy, information objectivity, user consent, and right-to-correct as our core ethical values. The potential ethical issues are related to the specific application context and we adopt a series of ethical protection measures throughout the design, development, and evaluation of AdSEE. During the collection of our QQ-AD dataset, we check the copyright licenses for each image and only select images with appropriate licenses that allow commercial use and free modification. We do not publish our QQ-AD dataset to ensure that copyright licenses are not violated and data privacy is protected. As a normal process, the platform advertisement censoring team censors every image on the platform for legal compliance and ethical control, which also includes all original and edited images used in the online experiments. The users participating in the online experiments are beta testers and internal employees who provided consent to opt into the beta testing program. The users have the option to provide feedback or opt out of the program at any time.

\section{Conclusion}
\label{sec:conclusion}
We present the AdSEE system which aims at finding out whether online advertisement visibility and attractiveness 
can be affected by semantic edits to human facial features in the ad cover images. Specifically, we design a CRP module to predict the click rate of an \textit{ad} based on the face latent embeddings offered by a StyleGAN-based encoder in addition to traditional image and textual embeddings. We further design a GADE module to efficiently search for optimal editing directions and coefficients using a genetic algorithm. Based on analyzing the introduced QQ-AD dataset, we identify semantic edit directions that are key to popularity enhancement. From the analysis,
we observe that a face oriented slightly downward, a smiling face, and a face with more feminine features are more attractive to users. Evaluation results on two offline datasets and online A/B tests demonstrate the existence of correlation between style editing and click rates in online \textit{ads}.

\clearpage
\onecolumn 
\begin{multicols}{2}
\bibliographystyle{ACM-Reference-Format}
\bibliography{sample-base}
\end{multicols}

\clearpage
\twocolumn
\appendix

\section{Base Recommender Models}
\label{sec:rec_models_compare}

To select the best-performing base recommender model for our task, we compare the performances among many SOTA models using our proposed set of features and on our dataset. For these experiments, we use the same set of features and the same experiment settings for each base recommender model for a fair comparison.

We briefly introduce the most important base recommender models in our comparison due to the vast amount of models compared. \citeauthor{rendle2010factorization} propose the Factorization Machine (FM)~\cite{rendle2010factorization} model to learn the first- and second-order interactions of features. 
To model the interactions of both the sparse and dense features, 
Wide~\&~Deep~\cite{wideanddeep} uses DNN to extract dense features and adopts a Logistic Regression (LR) model to learn the interactions between the dense and the sparse features. However, the Wide~\&~Deep~\cite{wideanddeep} model requires manual feature engineering for the sparse features, which needs domain expertise. 
To alleviate this downside, \citeauthor{deepfm} propose the DeepFM~\cite{deepfm} model to learn the first-order and high-order interactions automatically with an FM module and a DNN module, respectively.
Recently, the AutoInt~\cite{autoint} model was proposed which utilizes state-of-the-art deep learning techniques including attention mechanism~\cite{vaswani2017attention}, and residual connections~\cite{he2016deep} to learn both the first-order and high-order interactions automatically. 

Table~\ref{tab:rec-models-baseline} summarizes the performances of the different base recommender models on the QQ-AD dataset. Each model shares the same set of features described in Section~\ref{subsec:method_ctr} to ensure a fair comparison, i.e. ad category, multi-hot class label, one-hot face count, latent face representation, cover image embedding, and text embedding. We can clearly see that AutoInt~\cite{autoint} model significantly outperforms all the other baselines on all the evaluation metrics. The high performance of the AutoInt base recommender model is likely due to its adoption of a powerful multi-head self-attentive neural network with residual connections to model both the low-order and high-order feature interactions. 
These experiments also demonstrate that our method and features used for CR prediction are scalable to many models of various sizes and different designs. Furthermore, we repeat this base recommend model comparison on the CreativeRanking dataset and found the AutoInt model also provides the most robust performance among the compared models.

\begin{table}[htbp]
\centering
\caption{Comparing the CRP predictor with our proposed combined set of features on different base recommender models on the QQ-AD dataset.}
\label{tab:rec-models-baseline}
\begin{tabular}{l|cccc}
\hline
Model &
  MAE $\downarrow$ &
  \begin{tabular}[c]{@{}c@{}}Spearman's\\ rho $\uparrow$\end{tabular} &
  \begin{tabular}[c]{@{}c@{}}Kendall's\\ tau $\uparrow$\end{tabular} &
  \begin{tabular}[c]{@{}c@{}}Pearson's\\ R $\uparrow$\end{tabular} \\ \hline
DeepFM~\cite{deepfm}                  & 0.0263          & 0.5006         & 0.3526        & 0.5970      \\
CCPM~\cite{CCPM}                      & 0.0269          & 0.4718         & 0.3301        & 0.5651      \\
PNN~\cite{PNN}                        & 0.0264          & 0.4865         & 0.3414        & 0.5849      \\
WDL~\cite{wideanddeep}                & 0.0264          & 0.4973         & 0.3502        & 0.5949      \\
MLR~\cite{MLR}                        & 0.0265          & 0.4818         & 0.3380        & 0.5938      \\
NFM~\cite{NFM}                        & 0.0264          & 0.4994         & 0.3518        & 0.5982      \\
AFM~\cite{AFM}                        & 0.0270          & 0.4661         & 0.3260        & 0.5630      \\
DCNMix~\cite{DCNv2}                   & 0.0267          & 0.4863         & 0.3414        & 0.5893      \\
xDeepFM~\cite{xdeepfm}                & 0.0264          & 0.5078         & 0.3570        & 0.5918      \\
\textbf{AutoInt~\cite{autoint}}       & \textbf{0.0262} & \textbf{0.5122} & \textbf{0.3609} & \textbf{0.6113} \\
ONN~\cite{ONN}                        & 0.0264          & 0.4893         & 0.3439        & 0.5883      \\
FiBiNET~\cite{fibinet}                & \textbf{0.0262} & 0.4964         & 0.3499        & 0.6059      \\
IFM~\cite{IFM}                        & 0.0269          & 0.4818         & 0.3369        & 0.5673      \\
DIFM~\cite{DIFM}                      & 0.0268          & 0.4888         & 0.3418        & 0.5692      \\
AFN~\cite{AFN}                        & 0.0268          & 0.4815         & 0.3363        & 0.5626      \\ \hline
\end{tabular}
\end{table}

\section{Qualitative Study}
\label{sec:case_study}

\subsection{Semantic Editing Directions}
\label{subsec:edit_ditections}

\begin{figure}[htbp]
  \centering
  \includegraphics[width=2.8in]{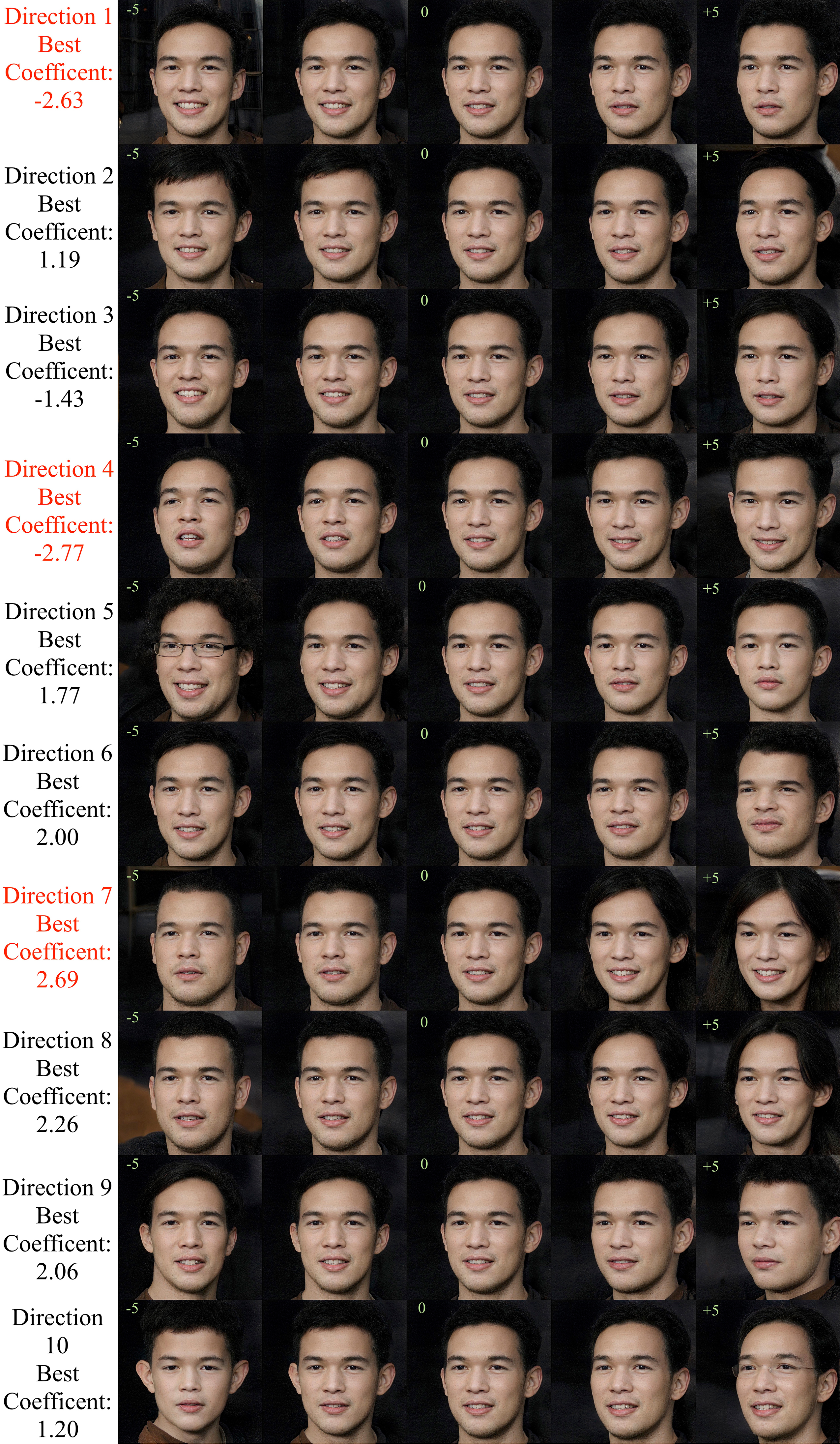}
  \caption{
  Case study analysis of edit directions 1 to 10.}
  \Description{Case Study Analysis of 10 Different Edit Directions.}
  \label{fig:case_edit_direction}
\end{figure}

In Figure \ref{fig:edit_compare_bar} from Section~\ref{sec:results}, we observe that editing on directions $n_4$, $n_7$, $n_1$ results in the largest increase in $\hat{CR}$. That is, these directions have the largest impact on the attractiveness of an \textit{ad} among the other editing directions. We further analyze the semantics of the editing directions in Figure \ref{fig:case_edit_direction}.
We visualize each editing direction by generating images from a range of editing intensity coefficients. Each row in the figure corresponds to one editing direction, and each column corresponds to a particular editing intensity value in the range of ${-5,-2.5,0,2.5,5}$ from left to right. 
We can see that for direction $n_4$, which corresponds to the vertical orientation of the face, the best average editing coefficient value found by the AdSEE model is -2.77 which means a face slightly facing downward is found to be more attractive. 
Similarly, for direction $n_7$, which corresponds to the gender of the face, the best average editing coefficient value found by the AdSEE model is 2.26 which means a face with more feminine features is more attractive. 
With editing direction $n_1$, which corresponds to the smilingness of the face, the best average editing coefficient value found by AdSEE is -2.63 which shows that a person with a smiling face is more attractive.

\subsection{AdSEE Edited Images}
\label{subsec:image_examples}

\begin{figure}[htbp]
  \centering
  \includegraphics[width=3.2in]{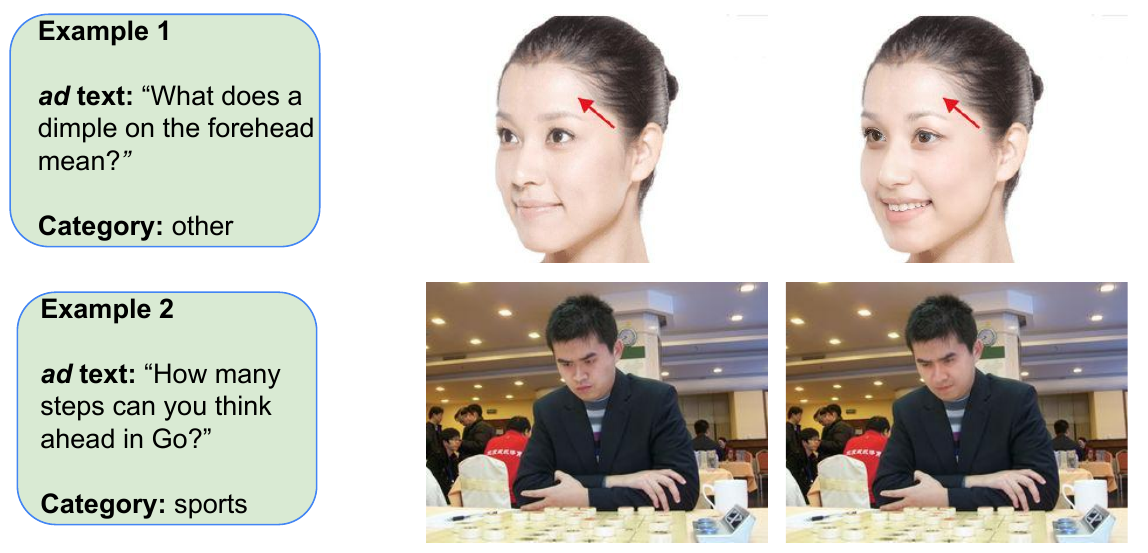}
  \caption{
  Examples of $ad$s enhanced by AdSEE where we show the $ad$ category, text, and cover image. Left: Original cover image, Right: Enhanced cover image.}
  \Description{Examples of ads enhanced by AdSEE where we show the ad category, text, and cover image. Left: Original cover image, Right: Enhanced cover image.}
  \label{fig:example_image}
\end{figure}

In this section, we qualitatively evaluate AdSEE by showing examples of AdSEE-enhanced images. Figure~\ref{fig:example_image} shows some examples of the enhanced \textit{ads} by the AdSEE model. The two examples are from two different \textit{ad} categories, i.e., others and sports. Nevertheless, the AdSEE model consistently chooses to enhance the attractiveness of the face by making it smile. In addition, the eyes in Examples 1 and 2 are edited to look downwards. These observations match our analysis of the average editing coefficient, where smilingness and vertical face orientation are attractive editing directions.

\section{Reproducibility}
\label{sec:reproducibility}

\textbf{Environment.}
We open source the implementation of AdSEE\footnote{Code available at \url{https://github.com/LiyaoJiang1998/adsee}.} so our method can be easily studied, reproduced, and extended. For all the experiments, we implement our model with PyTorch 1.7.0~\cite{PyTorch2} in Python 3.7.16 environment and train on a Tesla P40 GPU with a memory size of 24 GB. We also try our system on an RTX 2080Ti GPU with 11GB of memory, which can still handle our entire AdSEE system efficiently when tuning down the batch size hyperparameters. We provide the virtual environment and dependency setup script in our code repository for reproducibility.

\textbf{Pre-trained Models.}
Besides the important e4e~\cite{e4e} encoder model, the StyleGAN2-FFHQ~\cite{stylegan2} generator, and the SOLO instance segmentation model~\cite{solo, solov2} described in Section~\ref{sec:method}, we enumerate all the pre-trained models in our system. We adopt the pre-trained Bert-Chinese \cite{bert} model to extract the 768-dimensional text embedding of the \textit{ad} query texts as a dense feature. Within the Face Segmentation Module, we utilize the Dlib~\cite{dlib} face alignment model to extract aligned human faces. As for the Image Embedding Model, we use the Tencent internal Sougou Image Embedding Model and multi-label classification model described in Section~\ref{sec:method} for experiments on QQ-AD dataset. In addition, we adopt the publicly available ResNet-18~\cite{he2016deep} model as the image embedding model for experiments on the public CreativeRanking~\cite{wang2021hybrid} dataset. 
For the CRP-NIMA baseline model, we use the NIMA image quality assessment model \cite{talebi2018nima} pre-trained on the AVA \cite{gu2018ava} dataset to obtain the \textit{ad} cover image quality score mean and standard-deviation for all cover images in the QQ-AD dataset.

\textbf{Hyper-parameters and Implementation Details.}
For all of the base recommender models including AutoInt, we adopt the implementations from the DeepCTR~\cite{shen2017deepctr} library and use the default hyperparameters of each model. For both of the QQ-AD and CreativeRanking~\cite{wang2021hybrid}, we use the train set to train the model and use the validation set to tune the hyper-parameters, select features and determine early stop, and evaluate the performance on the test set. For the CRP model training on both datasets, we find a learning rate of $1e-4$ performs well and we use a batch size of 256. For the CRP model trained on the QQ-AD dataset and CreativeRanking dataset, we train them for 37 epochs and 18 epochs respectively.

For the GADE module, we use the following settings for experiments on the QQ-AD dataset. In Algorithm~\ref{alg:algorithm}, we set the \textit{PopulationSize} to 75 and set the \textit{NumGenerations} to 20. In the \textit{Parent Selection} step, we select 10 genotypes as parents by performing the rank selection method. In the \textit{Crossover} step, the parents in the mating pool will create 65 off-springs using the uniform crossover operation. Then, the 10 parents are combined with the 65 offspring to form a new population of 75 genotypes. Next, we randomly select 20\% of the 75 genotypes to mutate in the \textit{Mutation} step. For each genotype selected for mutation, we randomly change one of its genes by perturbing its value by a value in the range of $[-0.1,0.1]$. The search space for each gene value is limited to a value between the range of $[-3,3]$ with a step of 0.1. The gene values are randomly initialized to a value in the range of $[-1,1]$. In each genotype, we have 20 genes that correspond to the top 20 editing directions found by SeFa. 

On the CreativeRanking~\cite{wang2021hybrid} dataset, we use a slightly different set of settings for the GADE module that is suitable for this dataset. We use a \textit{PopulationSize} of 30 and set the \textit{NumGenerations} to 5. In the \textit{Parent Selection} step, we select 10 genotypes as parents by performing the rank selection method. In the \textit{Crossover} step, the parents in the mating pool will create 20 off-springs using the uniform crossover operation. Then, the 10 parents are combined with the 20 offspring to form a new population of 30 genotypes. Next, we randomly select 20\% of the 30 genotypes to mutate in the \textit{Mutation} step. For each genotype selected for mutation, we randomly change one of its genes by perturbing its value by a value in the range of $[-0.01,0.01]$. The search space for each gene value is limited to a value between the range of $[-1.5,1.5]$ with a step of 0.01. The gene values are randomly initialized to a value in the range of $[-0.1,0.1]$. In each genotype, we have 20 genes that correspond to the top 20 editing directions found by SeFa. 

For the operation used to convert face latent codes to a fixed length, we compare four operations including max-pooling, average-pooling, aggregation and concatenation with padding to a fixed length. We found max-pooling to be the best performer on both datasets and use the max-pooling operation throughout our experiments. In the implementation, we apply standardization to the click rate label and we predict the standardized click rate. Even more detailed implementation-related settings and their values can be found in the code.

\end{document}